%% file: iclr2023_conference.tex
\documentclass{article} 

\input{math_commands.tex}

\usepackage{booktabs}       
\usepackage{multirow}
\usepackage{listings}
\usepackage[left=1.1in,right=1.1in,top=1.3in,bottom=1.3in]{geometry}
\usepackage{xcolor}
\usepackage{natbib}
\usepackage{hyperref}
\usepackage{url}
\usepackage{graphicx}
\usepackage{subcaption}
\usepackage{wrapfig}
\usepackage{authblk}

\definecolor{codegreen}{rgb}{0,0.6,0}
\definecolor{codegray}{rgb}{0.5,0.5,0.5}
\definecolor{codepurple}{rgb}{0.58,0,0.82}
\definecolor{backcolour}{rgb}{0.95,0.95,0.92}

\lstdefinestyle{mystyle}{
  backgroundcolor=\color{backcolour}, commentstyle=\color{codegreen},
  keywordstyle=\color{magenta},
  numberstyle=\tiny\color{codegray},
  stringstyle=\color{codepurple},
  basicstyle=\ttfamily\footnotesize,
  breakatwhitespace=false,         
  breaklines=true,                 
  captionpos=b,                    
  keepspaces=true,                 
  numbers=left,                    
  numbersep=5pt,                  
  showspaces=false,                
  showstringspaces=false,
  showtabs=false,                  
  tabsize=2
}

\title{Extensible Proxy for Efficient NAS}

\author[1]{Yuhong Li}
\author[2]{Jiajie Li}
\author[3]{Cong Hao}
\author[4]{Pan Li}
\author[2]{Jinjun Xiong}
\author[1]{Deming Chen}
\affil[1]{University of Illinois Urbana-Champaign}
\affil[2]{University at Buffalo}
\affil[3]{Georgia Institute of Technology}
\affil[4]{Purdue University}

\begin{document}

\maketitle

\begin{abstract}

Neural Architecture Search (NAS) has become a de facto approach in the recent trend of AutoML to design deep neural networks (DNNs). 
Efficient or near-zero-cost NAS proxies are further proposed to address the demanding computational issues of NAS, where each candidate architecture network only requires one iteration of backpropagation. The values obtained from the proxies are considered the predictions of architecture performance on downstream tasks.
However, two significant drawbacks hinder the extended usage of Efficient NAS proxies.
(1) \textnormal{Efficient proxies are not adaptive to various search spaces.}
(2) \textnormal{Efficient proxies are not extensible to multi-modality downstream tasks.}
Based on the observations, we design a \underline{E}xtensible proxy (Eproxy) that utilizes self-supervised, few-shot training (i.e., 10 iterations of backpropagation) which yields near-zero costs. The key component that makes Eproxy efficient is an untrainable convolution layer termed barrier layer that add the non-linearities to the optimization spaces so that the Eproxy can discriminate the performance of architectures in the early stage. 
Furthermore, to make Eproxy adaptive to different downstream tasks/search spaces, we propose a \underline{D}iscrete \underline{P}roxy \underline{S}earch (DPS) to find the optimized training settings for Eproxy with only handful of benchmarked architectures on the target tasks. 
Our extensive experiments confirm the effectiveness of both Eproxy and Eproxy+DPS. On NAS-Bench-101 ($\sim$423k architectures), Eproxy achieves 0.65 as the spearman $\rho$. In contrast, the previous best zero-cost method achieves 0.45.
On NDS-ImageNet search spaces, Eproxy+DPS delivers 0.73 Spearman $\rho$ average ranking correlation while the previous efficient proxy only achieves 0.47. 
On NAS-Bench-Trans-Micro search space (7 tasks), Eproxy+DPS delivers comparable performance with early stop methods which requires 660 GPU hours per task. For the end-to-end task such as DARTS-ImageNet-1k, our method delivers better results compared to NAS performed on CIFAR-10 while only requiring a GPU hour with a single batch of CIFAR-10 images. Code is available at \url{https://github.com/leeyeehoo/GenNAS-Zero}.

\end{abstract}

\section{Introduction}
As deep neural networks (DNNs) find uses in a wide range of applications, such as computer vision~\citep{krizhevsky2012imagenet,simonyan2014very,he2016deep,redmon2016you} and natural language processing~\citep{vaswani2017attention,schuster1997bidirectional,hochreiter1997long,wu2020lite,devlin2018bert}, Neural Architecture Search (NAS)~\citep{zoph2018learning,real2019regularized,tan2019mnasnet,cai2019once,liu2018darts} has become an increasingly important technique to automate the design of neural architectures for different tasks~\citep{weng2019unet,wang2020fcos,liu2022pvnas,gong2019autogan}. Recent progress in NAS has demonstrated superior results, surpassing those of human designs~\citep{zoph2018learning,wu2019fbnet,tan2019mnasnet}. However, one major hurdle for NAS is its high computation cost. 
For example, the seminal work of NAS~\citep{zoph2018learning} consumed 2000 GPU hours to obtain a high-quality DNN, a prohibitively high cost for many researchers. The high computation cost of NAS can be attributed to three major factors: (1) the large search space for candidate neural architectures, (2) the training of the various candidate neural architectures, and (3) the comparison of the solution quality of candidate neural architectures to guide the NAS search process. 
Subsequent NAS work has proposed various techniques to address the above issues, such as the limitation of the search space, the weight-sharing networks to reduce the training cost, and the efficient proxies for evaluating the candidate architectures. 

Out of the advancement, the latest efficient proxies showed that the quality of a neural architecture could be determined by a proxy metric computed within seconds without full training. Hence they are near zero cost. For example, ~\cite{mellor2021neural} delivered NASWOT to analyze the activations of an untrained network as a proxy and demonstrated some promising results.~\cite{abdelfattah2021zero} proposed various proxies, such as gradients normalization (Grad norm), one-shot pruning based on a saliency metric computed at initialization \citep{tanaka2020pruning,wang2020picking,lee2018snip}, and Fisher~\citep{theis2018faster} that performs channel pruning by removing activation channels that are estimated to have the most negligible effect on the loss.  
The above efficient proxies, however, have two significant drawbacks. 
First, the quality of efficient proxies varies widely for different search spaces. Most of the proxies deliver high correlations with search space limited in the small NAS Benchmarks, while in real-life applications, the size of search spaces are order-of-magnitude larger than the tabular benchmarks'. 
For example, Synflow achieves high ranking correlation on NAS-Bench-201~\cite{dong2020nasbench201} (0.74 Spearman $\rho$) but performan poorly on NAS-Bench-101~\citep{ying2019bench} (0.37) which is 27X larger than NAS-Bench-201.
(2) \textnormal{Efficient proxies are not extensible to multi-modality downstream tasks.} One concern is that they are implicitly designed for CIFAR-10-level classification tasks where proxies deliver promising prediction results. For example, NASWOT fails (0.03 $\rho$ as the average ranking correlation) on NAS-Bench-MR~\citep{ding2021learning} (9 real-world tasks). Moreover, most efficient proxies apply specified algorithms such as pruning to transform the weights of architectures into prediction values. The fixed algorithm limits the adaptability of proxies towards tasks beyond classification. Besides, some ZC proxies introduce unknown bias for their preferences for certain neural architectures~\citep{chen2021bench}. It has been shown empirically and theoretically~\citep{ning2021evaluating} that Synflow prefers large models. 


\begin{figure}[t]
\vspace{-8pt}
         \centering
         \includegraphics[width=\textwidth]{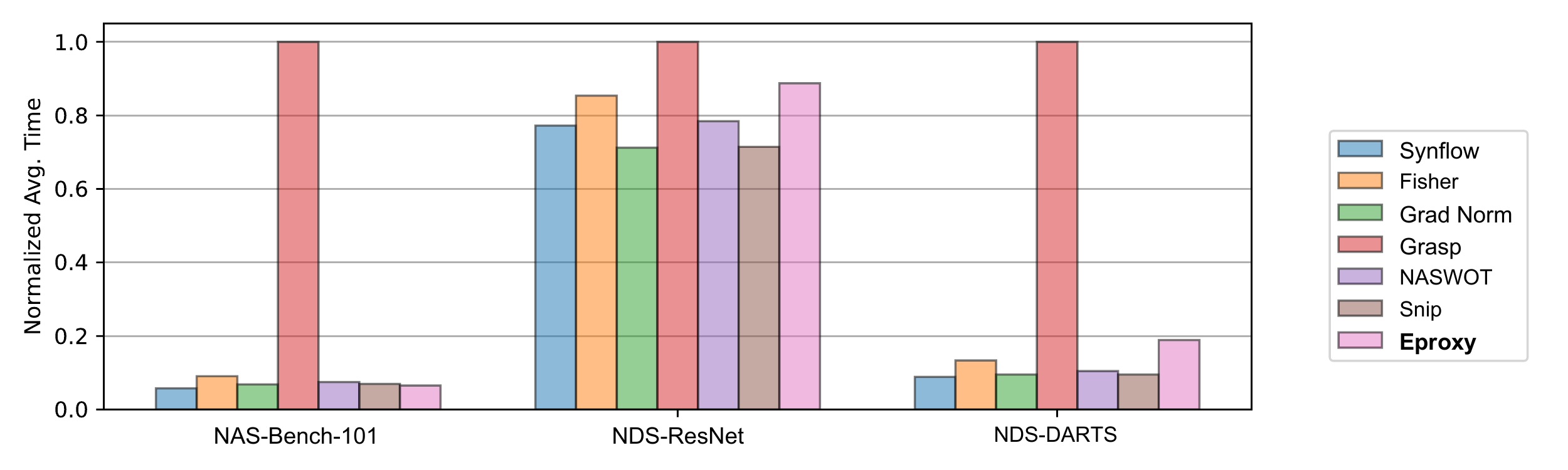}
           \vspace{-16pt}
         \caption{Comparison of Eproxy with six efficient proxies regarding evaluation speed on NAS-Bench-101, NDS-ResNet, and NDS-DARTS search spaces. The normalized average time is plotted. }
         \vspace{-16pt}
         \label{fig:timetrial}
\end{figure}

\begin{figure}[h]
\vspace{-8pt}
     \centering
     \begin{subfigure}[b]{0.49\textwidth}
         \centering
         \includegraphics[width=\textwidth]{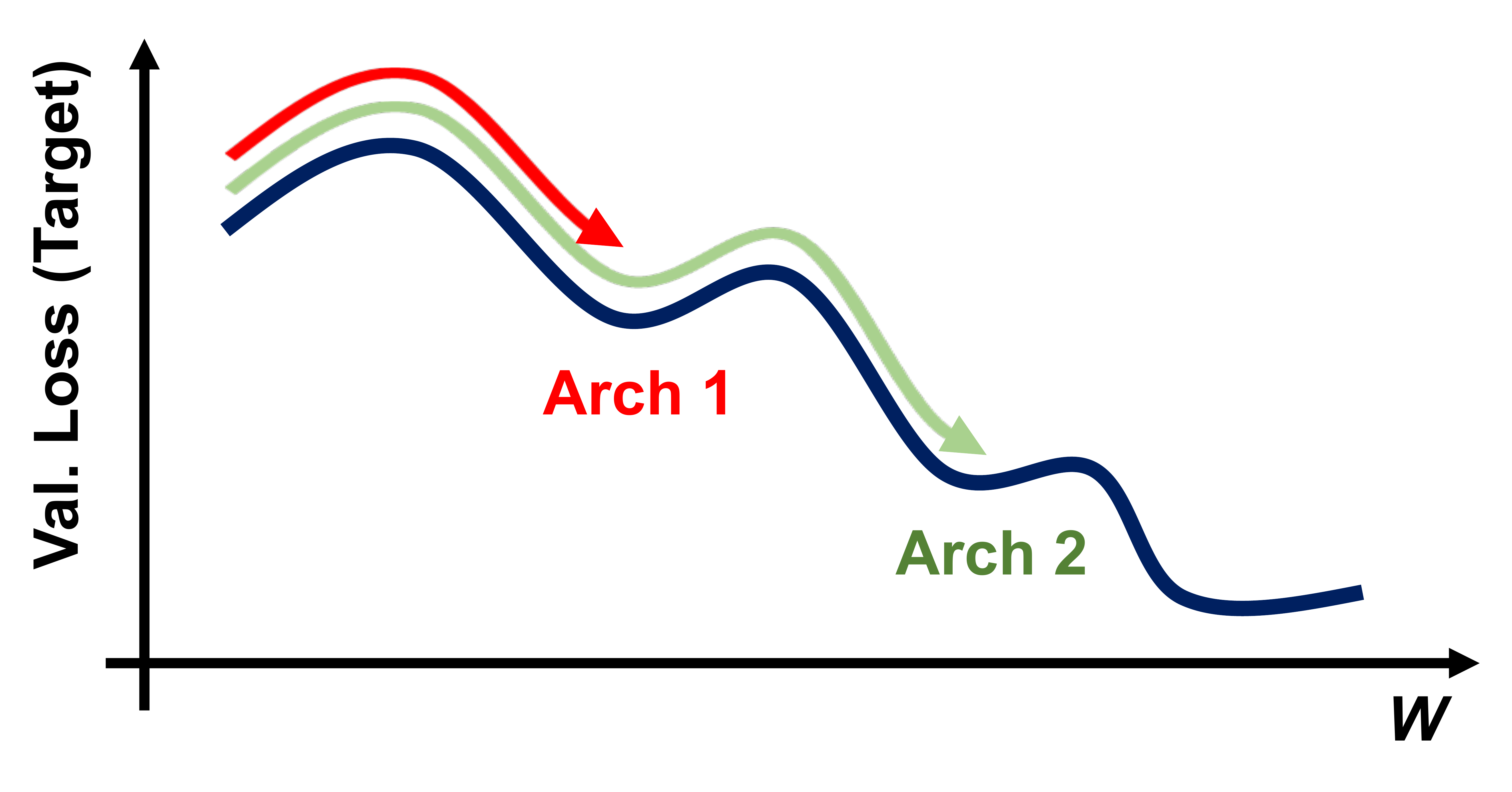}
     \end{subfigure}
     \hfill
     \begin{subfigure}[b]{0.49\textwidth}
         \centering
         \includegraphics[width=\textwidth]{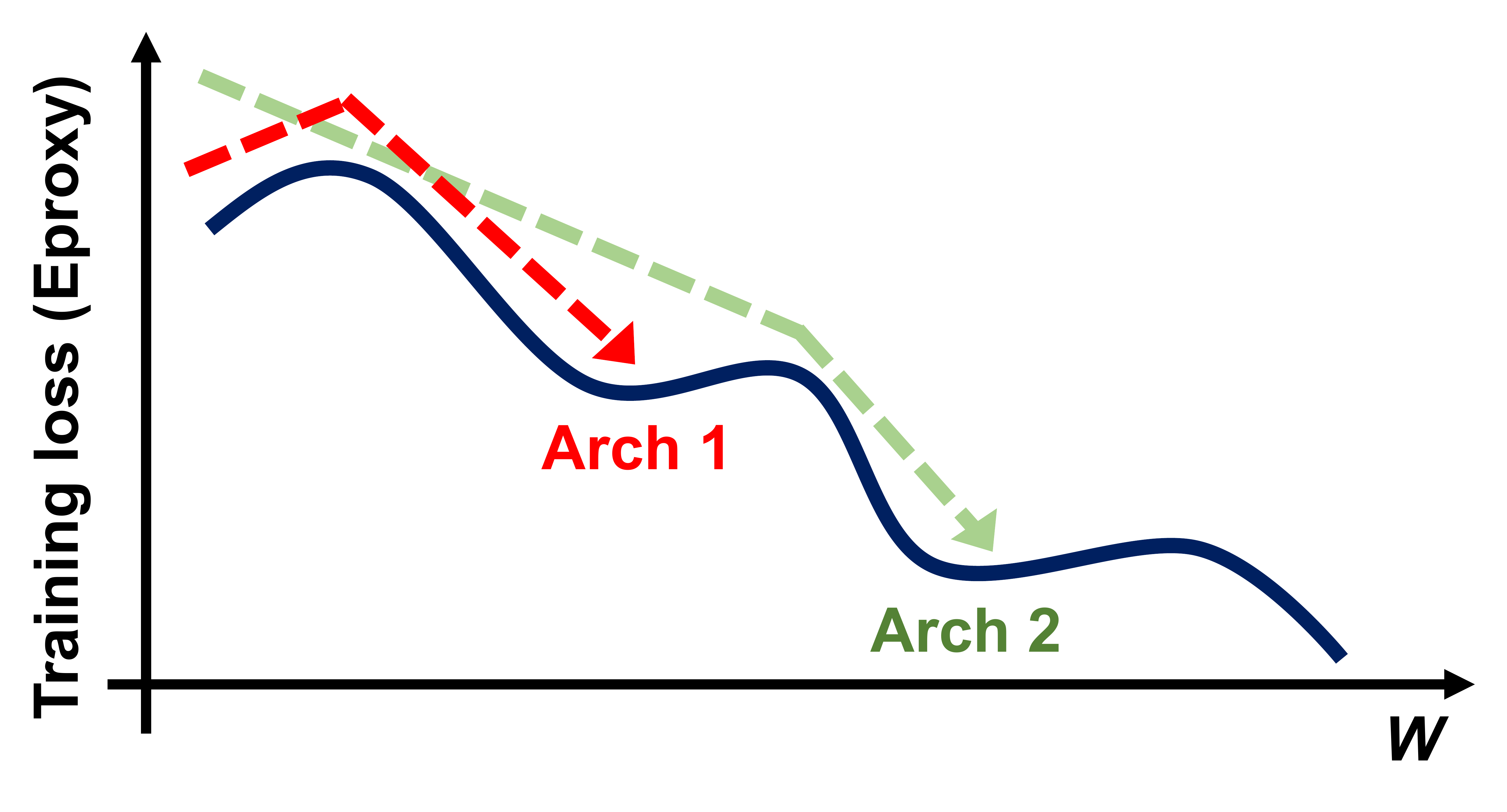}
     \end{subfigure}
        \caption{Illustration of the validation losses of two architectures on downstream task(left). A sophisticated few-shot proxy (right) can reflect the actual performance of architectures. }
        \label{fig:highlevel}
        \vspace{-8pt}
\end{figure}

\begin{figure}[h]
\vspace{-8pt}
     \centering
     \begin{subfigure}[b]{0.30\textwidth}
         \centering
         \includegraphics[width=\textwidth]{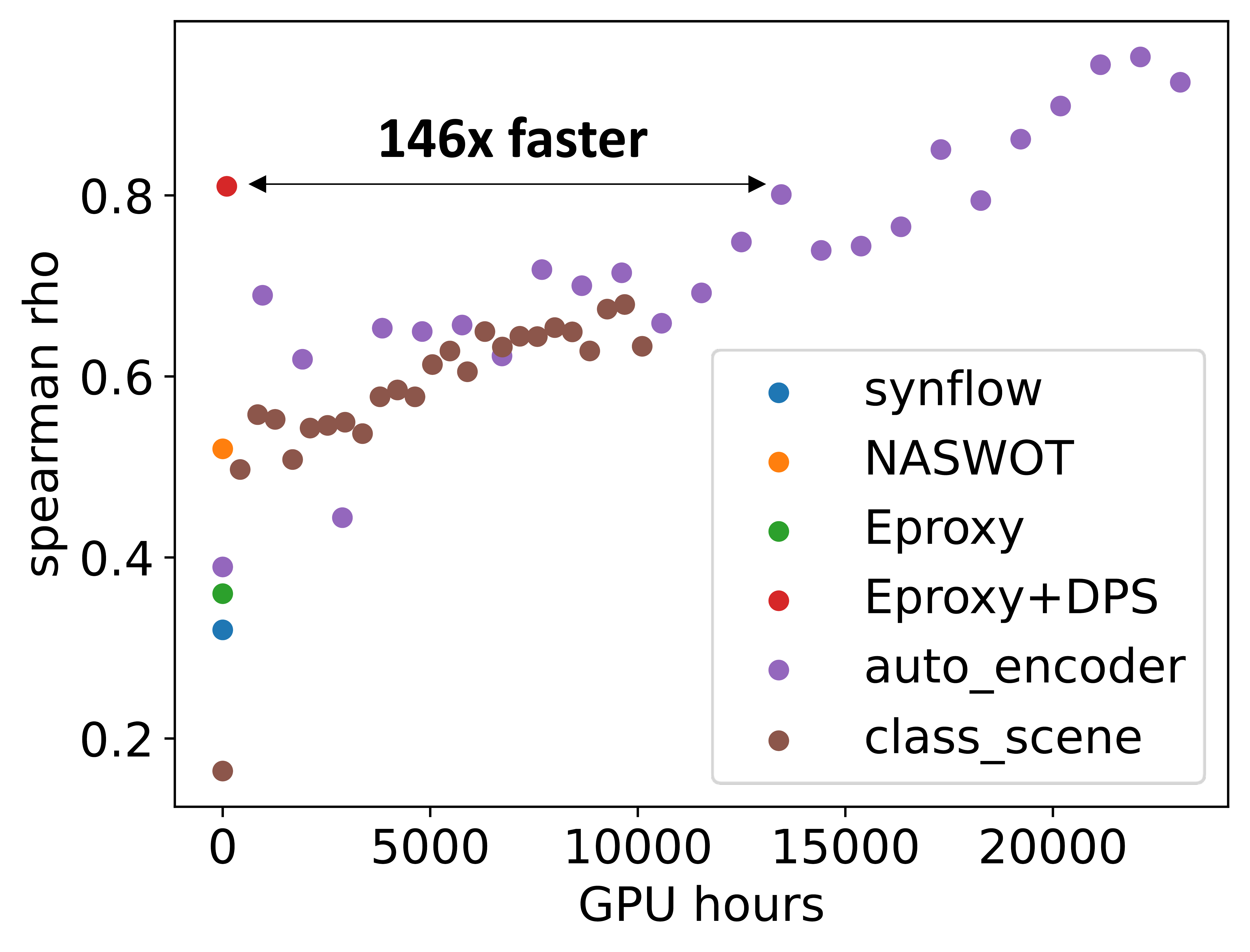}
         \caption{Autoencoder: Time vs Corr.}
         \label{fig:lr_gt}
     \end{subfigure}
     \hfill
     \begin{subfigure}[b]{0.33\textwidth}
         \centering
         \includegraphics[width=\textwidth]{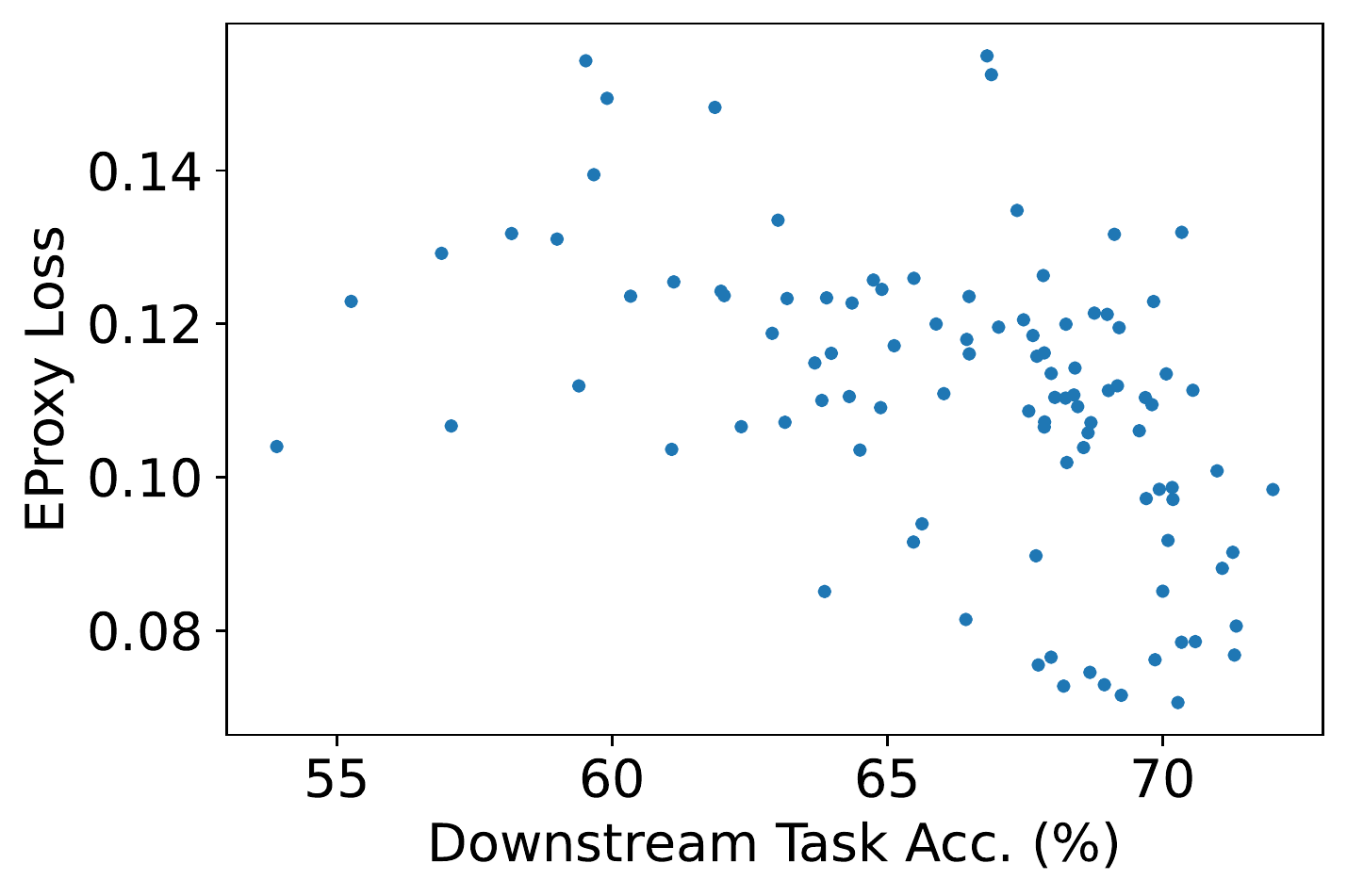}
         \caption{DARTS-ImgNet: Eproxy}
         \label{fig:darts eproxy}
     \end{subfigure}
     \begin{subfigure}[b]{0.33\textwidth}
         \centering
         \includegraphics[width=\textwidth]{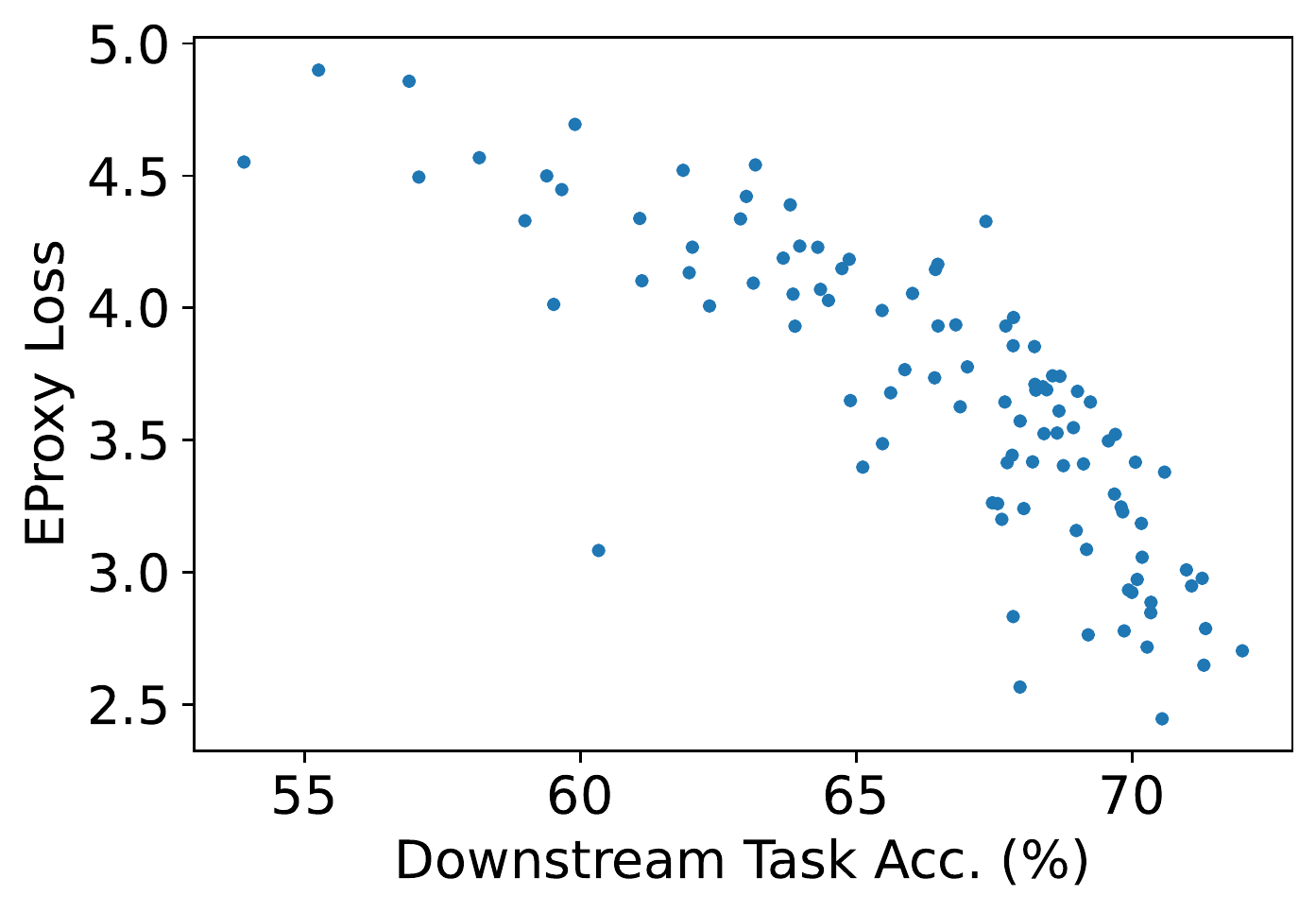}
         \caption{ DARTS-ImgNet: Eproxy+DPS}
         \label{fig:darts dps}
     \end{subfigure}
        \caption{\textbf{a:} Comparison with efficient proxies and early stopping methods on NAS-Bench-Trans-Micro Autoencoder task. It shows the effectiveness of DPS compared with early stopping methods on either the target task or a classification task when evaluating 4096 architectures. \textbf{b, c:} On NDS DARTS-\textbf{ImageNet} task, Eproxy and Eproxy+DPS (Searched on DARTS-CIFAR-10, transferred to ImageNet) achieve 0.51, 0.85 $\rho$ respectively. It shows DPS can find a search-space-aware Eproxy. }
        \label{fig:illustration}
        \vspace{-8pt}
\end{figure}

This work introduces a new efficient proxy termed \underline{E}xtensible proxy (Eproxy) from a different angle. Unlike previous efficient proxies,
Eproxy utilizes few-shot spatial-level regression on a set of image-label pairs (see Illustration in Fig.~\ref{fig:highlevel}). The labels are 2D synthetic features since spatial-level regression is more challenging than one-hot classification on a tiny dataset, i.e., a batch of image-label pairs as ~\cite{li2021generic} suggest. The key component of the Eproxy is the barrier layer. It takes the output of the architecture network into an untrained convolutional layer and performs the regression with the labels. Such a simple mechanism can significantly improve the performance of Eproxy to identify good architectures and bad ones when performing 10 iterations of backpropagation, i.e., near-zero cost. ($\sim$+0.57 $\rho$ ranking correlation on NAS-Bench-101.) We find that the barrier layer increases the complexity of the optimization space. Hence, poor-performance architectures are more difficult to optimize. (See Section ~\ref{sec:ablation nasbench101}).
Since Eproxy is a configurable few-shot trainer, we design a novel search space for Eproxy that includes various hyperparameters such as feature combinations, output channel numbers, and selection for barrier layers that makes Eproxy multi-modalities. We term the search method \underline{D}iscrete \underline{P}roxy \underline{S}earch (DPS) (The performance of DPS are shown in Fig.~\ref{fig:illustration}). Notably, besides the evaluation performance of a handful of architectures, DPS does not need to use any task-specific information (in our experiment, we only use a single batch of CIFAR-10~\citep{krizhevsky2009learning} images throughout all the experiments).
 
We summarize our contributions as follows:

\begin{itemize}
\item We propose an efficient proxy task with the barrier layer that utilizes a few-shot self-supervised regression. The task adopts only one batch of images in CIFAR-10-level dataset (not necessarily from the target training dataset). It uses the synthetic labels to evaluate architectures. Eproxy significantly speeds up the traditional early stopping evaluation process while maintaining the high ranking correlation.
\item We propose the downstream-task/search-space-aware proxy search algorithm with a proxy search space. We formulate the proxy task search as a discrete optimization problem with only a handful of architectures, such that the performance rankings of the networks on the ground-truth task and the proxy task should be consistent. The searched Eproxy can accurately evaluate the quality of network architectures and make Eproxy search-space/downstream-task aware.
\item We provide thorough experiments to evaluate the performance of Eproxy and Eproxy boosted by DPS on \textnormal{more than 30 search spaces/tasks}. We demonstrate that our methods have overall higher performance than existing efficient proxies in terms of all three factors: architecture ranking correlation score, top-10\%-architecture retrieve rate, and end-to-end NAS performance. Our solid experimental results can be further utilized and benefit the NAS community.
\end{itemize}

\section{Our Approaches}
In Sec.~\ref{sec: eproxy}, we introduce the Eproxy for efficient network evaluation; in Sec.~\ref{sec: dps}, we discuss how to find a downstream-task/search-space-aware Eproxy via Discrete Proxy Search.

\subsection{Extensible NAS Proxy}\label{sec: eproxy}

\begin{figure}
     \centering
     \includegraphics[width=0.8\textwidth]{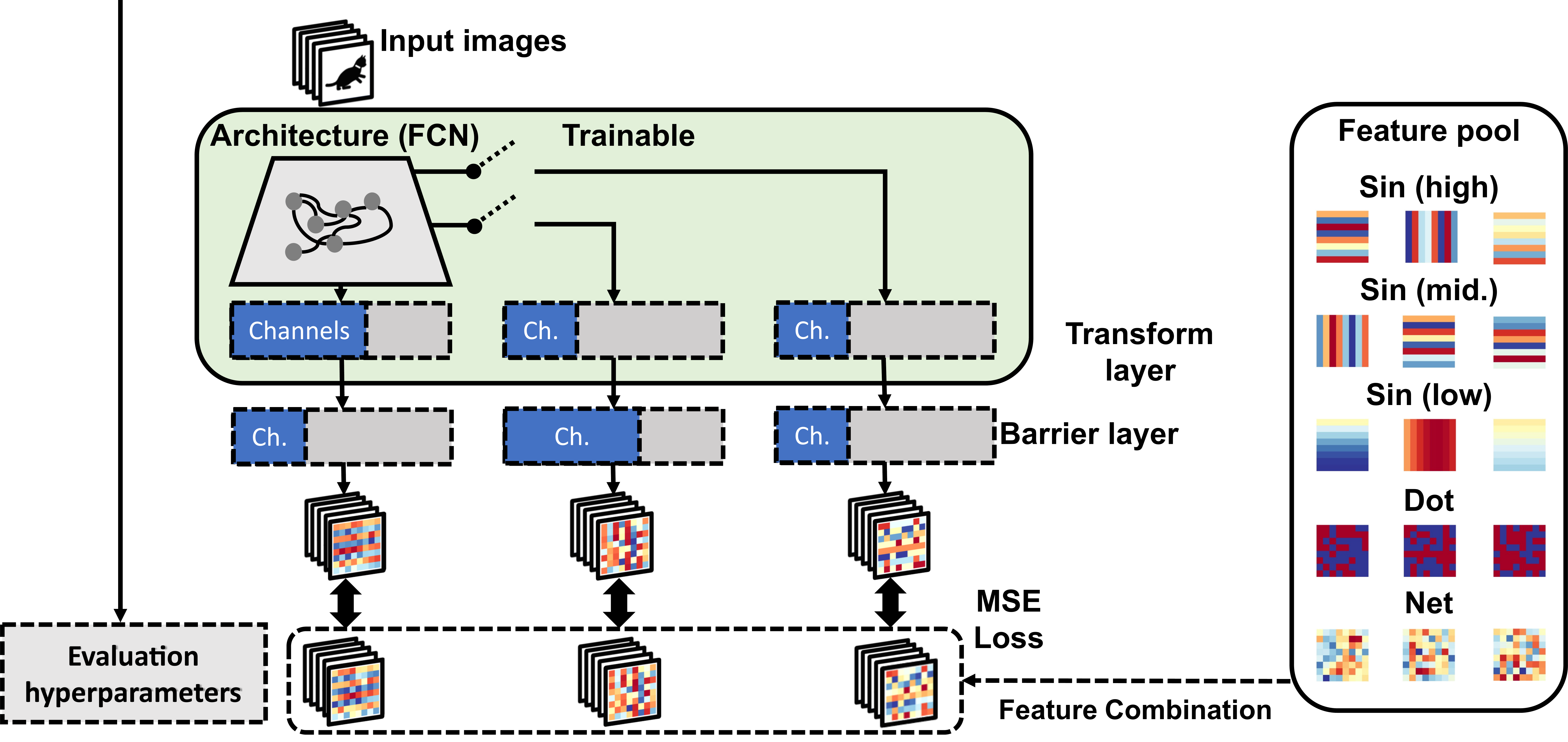}
        \caption{The design of Eproxy and the searchable components. Dotted line: The configurable components for Discrete Proxy Search. Green block: Trainable components. The configurability of Eproxy can be further utilized by DPS to target search spaces/downstream tasks.
        }
        \label{fig:pipeline}
\end{figure}

The Eproxy is designed for the architectures to learn the output of an untrained network on a set of image-label pairs (See Fig.~\ref{fig:pipeline}). 
We utilize the MSE-based training~\citep{li2021generic} with a large learning rate and limited backpropagation steps to make it as efficient as the existing near-zero-cost proxies. However, directly applying a few-shot regression task with a large learning rate leads to poor correlation based on our observations. To make the Eproxy architecture-performance-aware within a few iterations, we propose an untrained barrier layer to make the task more involved (See Section~\ref{sec:ablation nasbench101}).
The barrier layer is a randomly initialized convolution layer to the output of the trainable components. Our experiments show that adding such a layer can significantly improve the correlation between the predictions and the performance of neural architectures in the downstream tasks within a few back propagations (Sec~\ref{sec:ablation nasbench101}). To be more specific, the Eproxy training loss can be described as:

\begin{equation}
    \min_{w_a, w_t} ~\mathcal{L}_{MSE}(G(w_b, F(w_a, w_t, X)), Y)
\end{equation}

where the $X\in \mathbb{R}^{b\times c_{in} \times h_{in} \times w_{in}}$ is the a set of input images ($b$ is batch size; $c_{in}$ is number of input channels). 
$F$ is a fully convolutional neural network (FCN) with a transform layer (a convolutional layer) that transforms the $X$ to $F(\cdot) \in \mathbb{R}^{b\times c_{mid} \times h_{out} \times w_{out}}$. The FCN is usually obtained by utilizing the architecture without a task-specified head in the downstream tasks. For example, the classfier network with the classification (average pooling and linear layer) head removed.
$w_a$ and $w_t$ are the weights of architecture for evaluation and the weights of the transform layer (a convolution module) that project the output channels of the architecture to $c_{mid}$ which is the number of the transform layer's output channels. $G$ is the barrier layer, and $w_b$ is the weights. Note in the Eproxy without DPS, $Y\in \mathbb{R}^{b\times c_{out} \times h_{out} \times w_{out}}$ is the output of an untrained 6-layer FCN (Fig.~\ref{fig:pipeline}, `Net'). We interpret that Eproxy conducts a few-shot, tiny knowledge-distillation task from an untrained teacher network.

\subsection{Discrete Proxy Search}\label{sec: dps}

Since Eproxy provides abundant configurable hyperparameters and utilizes data-agnostic spatial labels, the different settings can be naturally adjusted for tasks/search spaces. Therefore, we propose a semi-supervised discrete proxy search to find a setting that can be suitable for the specific modality. As shown in Fig.~\ref{fig:pipeline}, the searchable configurations are provided as follows:
\begin{enumerate}
    \item Transform and barrier layer: Both layers can have kernel size selected from $\{1, 3, 7\}$, and the channel number $c_{mid}$ can be selected from 16 to 512 geometrically with 2 as a multiplier.
    \item Feature combination: a) Untrained FCN outputs. The experiment results show that an untrained network's output features can be powerful for evaluating architectures on numerous tasks/search spaces. b) Sine wave: we adopt the sine wave features with low/mid/high frequency along width/height. The insight is that good CNNs can learn different frequency signal~\citep{li2021generic, xu2019frequency}. c) Dot: By utilizing the Rademacher distribution, we generate the synthetic features with only $\pm 1$. The features attempt to simulate the spatial classification that is widely adopted in tasks such as detection~\citep{girshick2015fast}, segmentation~\citep{bertinetto2016fully}, tracking~\citep{bertinetto2016fully,li2018high}. For more details, please refer to the Appendix. The combined features can be multiplied by an augment coefficient selected from 0.5 to 2 with 0.5 as a step.
    \item Training hyperparameters: a) Learning rate: we adopt the SGD optimizer, and the learning rate can be selected from 0.5 to 1.5 with 0.1 as the step. b) Initialization: we adopt two initialization methods, Kaiming~\citep{he2015delving} and Xavier~\citep{glorot2010understanding} with either Gaussian or Uniform initialization (total 4 choices).
    \item Intermediate output evaluation: We provide the choices to force the network to learn the intermediate outputs from the layer before the first or second downsample layer. The motivation is that earlier stages of the network have different learning behaviors from the deeper stages ~\citep{alain2016understanding}. Thus, monitoring the early stages can give more flexibility for adapting Eproxy to different tasks.
    \item FLOPS: As works ~\citep{javaheripi2022litetransformersearch,wu2019fbnet,ning2021evaluating,colin2022adeeperlook} suggested that FLOPS is a good indicator for architecture performance. Hence we incorporate the FLOPS normalized by the largest architecture in the search space with the Eproxy loss as $\mathcal{L}\cdot(1+\alpha\cdot\textnormal{norm(FLOPS)})$. $\alpha$ can be selected from -0.5 to 0.5 with 0.1 steps.
\end{enumerate}

The total number of configuration combinations in the proxy search space is ~$5\textnormal{e}15$. We utilize the regularization evolutionary algorithm (REA)~\citep{real2019regularized} to conduct the exploration efficiently. First, we randomly sample a small subset of the neural architectures in the NAS search space and obtain their ground truth ranking on the target task or a highly correlated down-scaled task (for example, CIFAR-10 is considered a good proxy for ImageNet). We then evaluate these networks using Eproxy with different configurations and calculate the performance ranking correlation $\rho$ of the Eproxy and the target task, and the $\rho$ is the fitness function for REA.

\section{Experiments}

In this section, we perform the following evaluations for Eproxy and DPS. First, in Sec.~\ref{sec:ablation nasbench101}, we conduct the ablation study on NASBench-101~\citep{ying2019bench}, the first and yet the largest tabular NAS benchmark with over 423k CNN models and training statistics on CIFAR-10. We explain the mechanism behind the barrier layer with empirical results. Furthermore, we compared Eproxy and Eproxy boosted by DPS with existing efficient proxies.
Second, from Sec.~\ref{sec:nds} to Sec.~\ref{sec:nas-mb}, we use metrics including ranking correlation, top-10 architecture retrieve rate~\citep{dey2021ranking} to evaluate the proposed method on \textbf{NDS}~\citep{radosavovic2020designing} (11 search spaces on CIFAR-10, 8 search spaces on ImageNet), \textbf{NAS-Bench-Trans-Micro}~\cite{duan2021transnas} (7 tasks), and \textbf{NAS-Bench-MR}~\cite{ding2021learning} (9 tasks). 
Third, in Sec.~\ref{sec:dps}, we evaluate the end-to-end NAS on NAS-Bench-101/201. 
Moreover, we report the end-to-end search on the DARTS-ImageNet search space in Sec.~\ref{sec:imagenet}.

\begin{figure}[h]
\vspace{-8pt}
     \centering
     \begin{subfigure}[b]{0.49\textwidth}
         \centering
         \includegraphics[width=\textwidth]{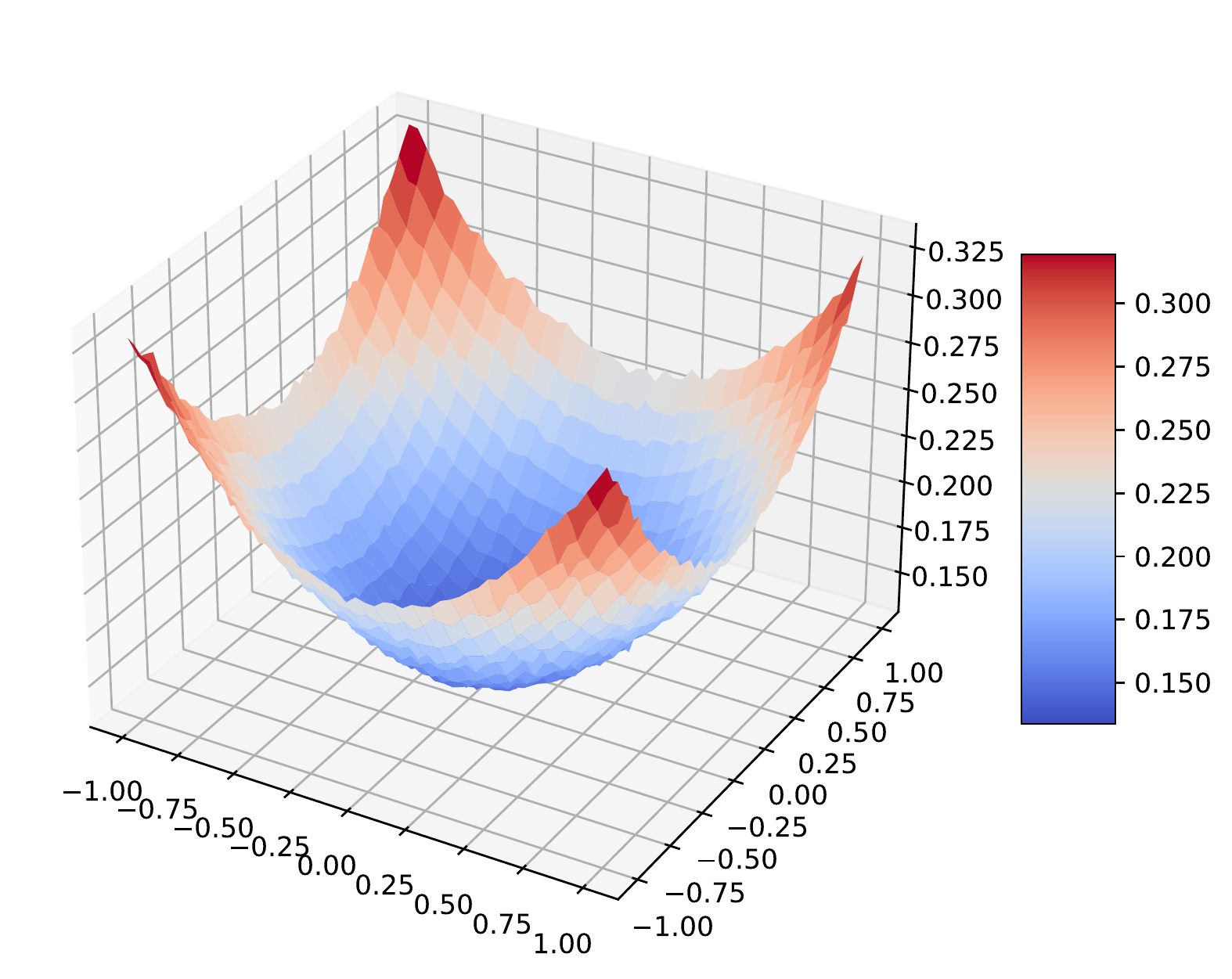}
         \caption{Regression without the barrier.}
         \label{fig:loss surface wo b}
     \end{subfigure}
     \hfill
     \begin{subfigure}[b]{0.49\textwidth}
         \centering
         \includegraphics[width=\textwidth]{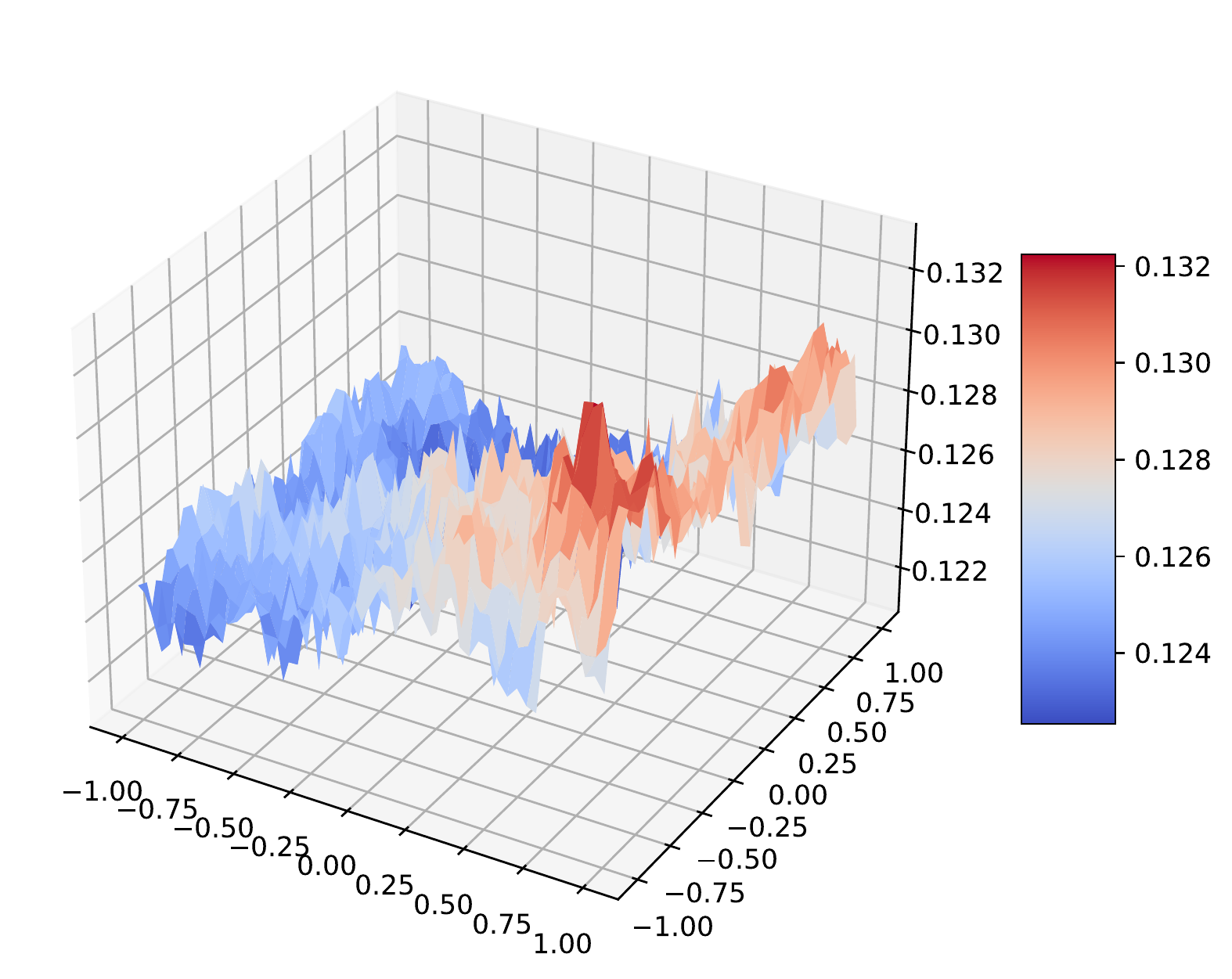}
         \caption{ Regression with the barrier (Eproxy).}
         \label{fig:loss surface w b}
     \end{subfigure}
        \caption{The loss surfaces of regression task with/without the barrier.}
        \label{fig:loss surface}
        \vspace{-8pt}
\end{figure}

\subsection{Ablation Study on NAS-Bench-101}\label{sec:ablation nasbench101}
\begin{table}[t]
  \centering
  \scriptsize
  \renewcommand{\arraystretch}{0.9}
\begin{tabular}{lccccccccc}
\toprule
 Loss & \multicolumn{3}{c}{MSE w/o Barrier} & \multicolumn{3}{c}{MSE w/ Barrier} \\\midrule
LR &    1   &   1e-1    &  1e-2    &     1   &   1e-1    &  1e-2        \\\midrule
10 iters$^{\textnormal{NZC}}$ &0.08&-0.22 &-0.19&\textbf{0.65}&0.46&0.09    \\
100 iters &0.07&0.67&0.76&0.65&\textbf{0.79}&0.79\\
200 iters &0.22&0.64&0.66&0.61&\textbf{0.83}&0.81    \\\bottomrule
\end{tabular}
  \caption{Ranking correlation (Spearman's $\rho$) analysis for different losses on NASBench-101. ``LR'' stands for learning rate; ``NZC'' stands for near-zero-cost. The results suggest that regression with barrier and large learning rate can achieve a high ranking correlation in 10 iterations near zero cost.}  \label{table:loss_ablation}

\end{table}
We study the effectiveness of our barrier layer in this section. We use the tool from \citep{li2018visualizing} to visualize the loss surface of an architecture selected randomly from NAS-Bench-101 on our few-shot regression task. Figure ~\ref{fig:loss surface} (a) shows the loss surface without the barrier has a good convexity, which indicates the task is simple, as we use a proxy task that contains very few samples (16 image-label pairs) for a shorter evaluation period. The simplicity of the proxy task gives us two potential problems that can affect the final results. (1) If a task is too simple, every model can perform similarly well. (2) When the optimization is easy, models can have similar performance at the early stage of training. As we observed, loss surfaces from different models have similar shapes without barriers, requiring us to use more training steps to see the difference between good and bad architectures. To mitigate these two problems, Eproxy added a barrier layer which is a random initialized linear/convolution layer with frozen weights. As shown in Figure 4 (b), the loss surface with the barrier has a noticeable non-convexity, which shows the increased complexity of the proxy task, and now it can better reflect the actual performance of architecture (See ~\ref{sec: moreloss} for more visualization). As the irregular shape of the loss surface varies widely from model to model, it helps us better distinguish the model performance at the early stage of training, allowing us to use fewer training steps to speed up the evaluation further.
The results in Table~\ref{table:loss_ablation} show that with the barrier layer, Eproxy can reach $\rho$ 0.65 in only 10 iterations with a learning rate of 1, and it also significantly improves the ranking correlation score with more training iterations.

Next, we sample 20 architectures from NAS-Bench-101 and evaluate DPS. We conduct DPS for 200 epochs, and the total run time is $\sim$20 mins on a single A6000 GPU.
In Table~\ref{table:nb101rho}, we report the network evaluation results in terms of Spearman's $\rho$ and top-10\% network coverage using the proxy task searched by DPS.
Eproxy significantly outperforms existing zero-cost proxies by a large margin. For example, Synflow, considered the stable proxy, achieves 0.45, NASWOT only achieves 0.38, Eproxy achieves 0.65 (without DPS), and Eproxy + DPS achieves 0.69. Regarding the top-10\% retrieve rate, Eproxy + DPS retrieves more architectures than DPS (38\% vs. 31\%). The results support the efficiency and effectiveness of DPS.
Meanwhile, Fig.~\ref{fig:timetrial} confirms that using Eproxy can achieve the same evaluation speed compared with other efficient proxies.

\begin{table}
\vspace{-8pt}
  \centering
  \scriptsize
  \renewcommand{\arraystretch}{0.9}
\begin{tabular}{lcccccccc}
\toprule
     &Grad norm&Snip&Grasp& Fisher & Synflow & NASWOT& Eproxy & Eproxy+DPS \\\midrule
$\rho$ &  0.20 &  0.16 & 0.45 & 0.26 & 0.37 & 0.40 & 0.65 & \textbf{0.69} \\
Top-10\% & 2\% & 3\% & 26\% & 3\% & 23\% & 29\% & 31\% & \textbf{38\%} \\\bottomrule
\end{tabular}
  \caption{Comparison with efficient proxies on NAS-Bench-101 using the Spearman $\rho$ and top-10\% retrieve rate. }  \label{table:nb101rho}

\end{table}
\subsection{NDS} \label{sec:nds}
~\cite{mellor2021neural} utilizes an interesting and practical dataset named Network Design Spaces (NDS), where the original paper aims to compare the search spaces themselves. The NDS is perfect for evaluating efficient proxies in more complex search spaces. For example, researchers benchmark 5,000 architectures on DARTS search space and over 20,000 on ResNet search space. We compared our method with existing zero-cost proxies on \textbf{11 search spaces} on \textbf{CIFAR-10} and \textbf{8 search spaces} on \textbf{ImageNet}~\cite{deng2009imagenet}.  We show the results in Table~\ref{table:ndsrho}. 
Compared to NASWOT~\citep{mellor2021neural}, Eproxy (without DPS) achieves on-a-par results on both CIFAR-10 and ImageNet search spaces. Boosted by DPS, Eproxy delivers significantly better results on target CIFAR-10 search spaces with 36\% and 52\% improvement on ranking correlation and top-10\% retrieve rate, respectively. Notably, Eproxy+DPS searched on CIFAR-10 with 20 architectures performs significantly better on \textbf{ImageNet} search spaces \underline{without any prior knowledge of the dataset}. Compared to NWT, Eproxy+DPS gains 30\% and 57\% on ranking correlation and top-10\% retrieve rate, respectively. The ImageNet experiment demonstrates the efficiency by utilizing the architectures trained on down-scaled dataset (CIFAR-10) for DPS.

\begin{table}
\small
  \setlength{\tabcolsep}{3pt}
  \renewcommand{\arraystretch}{0.8}
  \centering
  \scriptsize
\begin{tabular}{lcccccccccccc}
\toprule
CIFAR-10 & DARTS & DARTS-f & AMB & ENAS & ENAS-f & NASNet & PNAS & PNAS-f & Res & ResX-A &ResX-B &Avg. \\\midrule
\multirow{2}{*}{Synflow} & 0.42 & -0.14 & -0.10 & 0.18 & -0.30 &0.02  &0.25  & -0.26 & 0.21 &0.47  &\textbf{0.61} &0.12 \\
                  & 9\% & 5\% & 3\% & 6\% & 2\% & 7\% & 9\% & 4\% & 4\% & 25\% & \textbf{29\%} & 9\% \\\cmidrule(lr){2-13}
\multirow{2}{*}{NASWOT} & \textbf{0.65} & 0.31 & 0.29 & \textbf{0.54} & 0.44 & 0.42 & 0.50 &0.13  & 0.29 & \textbf{0.64} & 0.57 &0.43 \\
                  & \textbf{29\%} &  8\% & 20\% & \textbf{31\%} & 28\% & 27\% & 24\% & 6\% & 7\% & \textbf{28\%} & 21\% & 21\% \\\cmidrule(lr){2-13}
\multirow{2}{*}{Eproxy} & 0.38 & 0.34 & 0.54 & 0.59 & {0.48}  & 0.56 & 0.22 & 0.24 & 0.51 &  0.47 & 0.19 & 0.41\\
                  & 12\% & 17\% & 13\% & 35\% & 31\% & 28\% & 4\% & {4\%} & {36\%} & 24\% & 10\% & 19\% \\\cmidrule(lr){2-13}
\multirow{2}{*}{Eproxy+DPS} & \textbf{0.72} & \textbf{0.39} & \textbf{0.56} & \textbf{0.63} & \textbf{0.47} & \textbf{0.54} & \textbf{0.60} & \textbf{0.48} & \textbf{0.56} &  \textbf{0.65} & {0.60} & \textbf{0.56}\\
                  & \textbf{33\%} & \textbf{19\%} & \textbf{29\%} & \textbf{36\%} & \textbf{30\%} & \textbf{32\%} & \textbf{35\%} & \textbf{28\%} & \textbf{36\%} & \textbf{32\%} & {19\%} & \textbf{29\%} \\\midrule
\end{tabular}
\begin{tabular}{lcccccccccc}
\midrule
  ImageNet& DARTS & DARTS-f & Amoeba & ENAS & NASNet & PNAS  & ResX-A &ResX-B &Avg. \\\midrule
\multirow{2}{*}{Synflow}  & 0.21 & -0.36 & -0.25 & 0.17 &0.01  &0.14  & 0.42 &0.31 & 0.08   \\
                  & 0\% & 4\% & 0\% & 9\% & 0\% & 9\% & 7\% &  13\%  & 6\% \\\cmidrule(lr){2-10}
\multirow{2}{*}{NASWOT}  & {0.66} & 0.20 & 0.42 & {0.69} & {0.51} & \textbf{0.61} & {0.73} &    {0.63} & 0.56  \\
                  & 16\% & 8\% & 33\% & \textbf{36\%} & \textbf{33\%} & 10\% & 30\% &  \textbf{38\%}& 26\%    \\\cmidrule(lr){2-10}
\multirow{2}{*}{Eproxy}  & 0.51 &  {0.31} & 0.66 & 0.58 & 0.56 & 0.36 & 0.73 &   0.70 & 0.55 \\
                   & 20\% &  {17\%} & 60\% & 33\% & 30\% & {33\%} & 55\% &  43\%  & 36\% \\\cmidrule(lr){2-10}
                  
\multirow{2}{*}{Eproxy+DPS$_\textnormal{T}$} & \textbf{0.85} &  \textbf{0.53}  & \textbf{0.66} & \textbf{0.79} & \textbf{0.85} & \textbf{0.60} & \textbf{0.83} &   \textbf{0.72} &\textbf{0.73}   \\
                   & \textbf{50\%} &  \textbf{28\%} & \textbf{60\%} & 33\% & {32\%} & \textbf{35\%} & \textbf{55\%} &  36\% & \textbf{41\%}   \\\bottomrule
\end{tabular}
  \caption{Comparison with efficient proxies on NDS search spaces. $_\textnormal{T}$ denotes the DPS is conducted on CIFAR-10 and directly transferred to \textbf{ImageNet}. }
  \label{table:ndsrho}
\end{table}

\begin{table}
  \centering
  \scriptsize
\begin{tabular}{lcccccccc}\toprule
                  & Cls. Scene & Cls Obj & Room Layout & Jigsaw & Seg & Normal & AE & Avg. \\\midrule
              Synflow   & 0.46/16\%&0.50/{16\%}&0.45/28\%&0.49/19\%&0.32/3\%&0.52/19\%&{0.52}/{34\%} & 0.47/19\% \\
                NASWOT  & 0.57/{21\%}&0.53/{21\%}&0.30/2\%&0.41/11\%&0.52/30\%&0.59/{30\%}&-0.02/2\%& 0.41/17\%\\\midrule
                Eproxy  & 0.15/14\%&0.45/{34\%}&0.06/8\%&0.17/33\%&0.36/46\%&0.25/38\%&0.61/\textbf{80\%} & 0.29/36\%\\
                Eproxy + DPS  & \textbf{0.70/30\%}&\textbf{0.56/44\%}&\textbf{{0.56}}/13\%&\textbf{{0.64}/45\%}&\textbf{{0.81}/53\%}&\textbf{{0.81}/63\%}&\textbf{0.80}/74\% & \textbf{0.69}/\textbf{46\%}\\\midrule
                ES$_{\sim\textnormal{ 660GPU hrs/task}}$  & {0.73/25\%}&{0.01/7\%}&{{0.15}}/7\%&{{0.74}/21\%}&{{0.39}/7\%}&{{0.65}/27\%}&{0.35}/11\% & 0.43/15\%\\\bottomrule
\end{tabular}
  \caption{Comparison with efficient proxies and the early stopping method on TransNAS-Bench-Micro. Eproxy+DPS outperforms efficient proxies and early stopping method.}
  \label{table:transrho}
\end{table}

\subsection{NAS-Bench-Trans-Micro} \label{sec:nas-trans-micro}
Previous experiments suggest that DPS can optimize Eproxy across different search spaces. We further evaluate Eproxy and DPS on NAS-Bench-Trans-Micro, a benchmark that contains 4096 architectures across \textbf{7 large tasks} from the \textbf{Taskonomy}~\cite{zamir2018taskonomy} dataset. The tasks include object classification, scene classification, unscrambling the image, and image upscaling. The search space is similar to NAS-Bench-201 but has 4 operator choices per edge instead of 6. We conduct the DPS on each task using only 20 architectures. We do not have any prior knowledge of the tasks besides the 20 architecture's ground truth performance since DPS only utilizes a batch of CIFAR-10 images as input. We compare our method with NASWOT, Synflow, and the early stopping method shown in Table~\ref{table:transrho}. Note that though Eproxy underperforms regarding the ranking correlation, it achieves an 89\% higher top-10\% retrieve rate compared to Synflow. It also tells that the global ranking correlation is not the golden metric for evaluating the performance of proxies since it merely reflects the difference of top architectures.
With the help of DPS, the average ranking correlation and top 10\% retrieve rate are significantly improved and substantially better than other methods. Compared to the early stopping method, DPS requires 7.6X less regarding GPU hours ($>$99\% time for obtaining the performance of 20 architectures while the DPS only takes ~0.5 GPU hour).


\begin{table}[h]
  \centering
  \scriptsize
  \renewcommand{\arraystretch}{0.8}
\begin{tabular}{lcccccccccc}
\toprule
                  & Cls-A & Cls-B & Cls-C & Cls-10c & Seg & Seg-4x & 3dDet & Video & Video-p &Avg. \\\midrule
\multirow{2}{*}{Synflow} &0.25&0.05&0.37&0.21&0.43&0.22&0.22&0.45&0.52 & 0.30\\
                  & 11\% & 14\% & 20\% & 15\% & 17\% & 9\% & 8\% & 18\% & 17\% & 14.3\% \\\cmidrule(lr){2-11}
\multirow{2}{*}{NASWOT} & {0.37}&-0.20&-0.15&-0.39&{0.50}&{0.38}&{0.48}&-0.36&-0.36 & 0.03\\
                  & 18\% & 4\% & 2\% & 0\% & 10\% & 8\% & 10\% & 1\% & 0\% & 6\% \\\cmidrule(lr){2-11}
\multirow{2}{*}{Eproxy} & 0.52 & {0.06} & 0.02 & 0.29 & 0.38 & {0.31} & 0.34 & 0.31 & {0.23} &0.27\\
                  & 18\% & 10\% & 10\% & 15\% & 17\% & 13\% & 23\% & 11\% & 11\% & 14\%\\\cmidrule(lr){2-11}
\multirow{2}{*}{Eproxy + DPS} & \textbf{0.57} & \textbf{0.53} & {0.30} & \textbf{0.48} & \textbf{0.60} & \textbf{0.51} & 0.39 & \textbf{0.65} & \textbf{0.59} & \textbf{0.51} \\
                  & {16\%} & \textbf{35\%} & {18\%} & \textbf{32\%} & \textbf{24\%} & \textbf{13\%} & \textbf{29\%} & \textbf{33\%} & \textbf{27\%} & \textbf{25\%}\\\cmidrule(lr){1-11}
Cls-C Full training & 0.29 & {0.51} & 1.0 & 0.53 & 0.21 & {0.35} & 0.17 & 0.35 & {0.37} & n/a\\
($\sim$4000GPU hrs)& 24\% & 26\% & 100\% & 34\% & 16\% & 26\% & 14\% & 22\% & 25\% & N/A\\\bottomrule
\end{tabular}
  \caption{Comparison with efficient proxies and Cls-C full training on NAS-Bench-MR. Eproxy+DPS is comparable with the full training on Cls-C task.}
  \label{table:mbrho}
  
\end{table}
\subsection{NAS-Bench-MR} \label{sec:nas-mb}
We try the Eproxy and DPS on a more complex search space, NAS-Bench-MR~\citep{ding2021learning}, with \textbf{9 high-resolution tasks} such as 3d detection, ImageNet-level classification, segmentation, and video recognition~\cite{deng2009imagenet,cordts2016cityscapes,Geiger2012CVPR,kuehne2011hmdb}. Randomly sampled $\sim$2,500 architectures are evaluated on the tasks from the entire search space. Each architecture is fully trained ($>$100 epochs) and follows a multi-resolution paradigm, where each network contains four stages. Each stage comprises modularized blocks (parallel and fusion modules). Hence, the benchmark is unprecedentedly complicated.
Our work is the first to investigate this benchmark with efficient proxies. We compared Eproxy and Eproxy+DPS with NASWOT, Synflow, and full training on Cls-C task (~4000GPU hrs~\footnote{\url{https://github.com/dingmyu/NCP}}). The results are shown in Table~\ref{table:mbrho}. 
Note that NASWOT, which performs well on NAS-Bench-Trans-Micro, delivers poor performance on most tasks, implying the inconsistent performance of current efficient proxies. Also, we observed that classification rankings are inconsistent with other tasks, such as segmentation and 3D detection. 
Our Eproxy+DPS experiments suggest that with a 20-architecture set, the ranking correlation and top-10\% retrieve rate are considerably improved (\textbf{+89\%/+78\%}). 

\subsection{End-to-end NAS with Eproxy}~\label{sec:dps}
We evaluate Eproxy and DPS on the end-to-end NAS tasks, aiming to find high-performance architectures within the search space.

\begin{table}
  \centering
  \scriptsize
  \renewcommand{\arraystretch}{0.8} 
\begin{tabular}{l|c|c|c|c|ccc|c|c|ccc}\toprule
          & RS    & NAO   & RE    & Semi  & \multicolumn{3}{c|}{WeakNAS} & Synflow  & NASWOT   & \multicolumn{3}{c}{Eproxy+DPS} \\ \midrule
Queries   & 2000  & 2000  & 2000  & 1000  & 200      & 150     & 100     & 0     & 0     & 150     & 60      & 0      \\\midrule
Test Acc. & 93.64 & 93.90 & 93.96 & 94.01 & 94.18    & 94.10   & 93.69   & 92.20 & 90.06 & \textbf{94.23}   & \textbf{93.92}   & \textbf{93.07} \\ \bottomrule
\end{tabular}
  \caption{Comparison with predictor-based methods and efficient proxies on NAS-Bench-101. Eproxy+DPS can find near-optimal architectures with lower queries.}
  \label{table:e2e nb101}
  
\end{table}

\begin{table}[h]

  \centering
  \scriptsize
\begin{tabular}{lcccccc}
\toprule
        & Random Search & Regularized Evolution & MCTS  & LaNAS & WeakNAS & Eproxy+DPS \\ \midrule
C10     & 7782.1        & 563.2         & 528.3 & 247.1 & 182.1   & \textbf{58.0 + 20}  \\ \midrule
C100    & 7621.2        & 438.2         & 405.4 & 187.5 & 78.4    & \textbf{13.7}$_\textbf{T}$      \\ \midrule
TinyImg & 7726.1        & 715.1         & 578.2 & 292.4 & 268.4   & \textbf{74.0}$_\textbf{T}$         \\ \bottomrule
\end{tabular}
\caption{Comparison with predictor-based methods on NAS-Bench-201 regarding the average queries required for retrieving the global optimal architectures. Eproxy+DPS uses substantially lower queries to find the global optimal architectures.}
  \label{table:end2end}
\end{table}

\begin{table}[h]
\scriptsize
  \centering
  \scriptsize
\begin{tabular}{lccccccc}
\toprule
\multirow{2}{*}{Method} & \multicolumn{2}{c}{Test Err. (\%)} & Params & FLOPS & Search Cost & Searched & Searched \\ \cline{2-3} 
                        & top-1          & top-5         &(M)&(M)&(GPU days)&Method&dataset
      \\ \midrule 
NASNet-A~\cite{zoph2018learning} &  26.0 &8.4&5.3&564&~2000&RL&CIFAR-10\\
AmoebaNet-C~\cite{real2019regularized} & 24.3 & 7.6 &  6.4 & 570 & 3150&evolution&CIFAR-10\\
PNAS~\cite{liu2018progressive}&25.8&8.1&5.1&588&225&SMBO&CIFAR-10\\
\midrule
DARTS(2nd order)~\cite{liu2018darts} &26.7& 8.7& 4.7& 574& 4.0& gradient-based&CIFAR-10\\

SNAS~\cite{xie2018snas}&27.3&9.2&4.3& 522& 1.5&gradient-based&CIFAR-10\\
GDAS~\cite{dong2019searching}&26.0&8.5&5.3&581&0.21&gradient-based&CIFAR-10\\
P-DARTS~\cite{chen2019progressive}&24.4&7.4 &4.9 &557 &0.3 &gradient-based &CIFAR-10\\
P-DARTS&24.7&7.5&5.1 &577 &0.3 &gradient-based &CIFAR-100\\
PC-DARTS~\cite{xu2019pc}&25.1 &7.8 &5.3 &586 &0.1 &gradient-based&CIFAR-10\\
TE-NAS~\cite{chen2021neural}&26.2 &8.3 &6.3&-& 0.05& training-free&CIFAR-10\\

\midrule
PC-DARTS&24.2 &7.3 &5.3 &597 &3.8 &gradient-based&ImageNet\\
ProxylessNAS~\cite{cai2018proxylessnas} &  24.9 &7.5 &7.1 &465 &8.3 &gradient-based&ImageNet\\
TE-NAS~\cite{chen2021neural}& 24.5  &7.5 &5.4 & 599&0.17&training-free&ImageNet\\
\midrule
\textbf{Eproxy} & 25.7&8.1& 4.9 &542& 0.02&evolution+proxy&CIFAR-10\\
\textbf{Eproxy+DPS}$_\textnormal{T}$ & 24.4&7.3&5.3&578&  0.06&evolution+proxy&CIFAR-10\\
\bottomrule
\end{tabular}
  \caption{Comparison with state-of-the-art NAS methods on ImageNet. $_\textnormal{T}$ stands for DPS is conducted in NDS search space and directly transferred to the target. Note Eproxy+DPS achieves the best results among NAS methods on CIFAR-10. 
  }
  \label{table:GenNAS_NAS_Imagenet}
\end{table}

On \textbf{NAS-Bench-101}, we utilize the Eproxy as the fitness function for Regularized Evolutionary (RE) algorithm. Our results are shown in Table~\ref{table:e2e nb101} compared with NAO ~\citep{luo2018neural}, Semi-NAS~\citep{luo2020semi}, WeakNAS~\citep{wu2021stronger}, Synflow~\citep{abdelfattah2021zero}, NASWOT~\citep{mellor2021neural}. Note that Eproxy, without any query (near-zero-cost) from the benchmark, can find architectures that are significantly better than current SoTA efficient proxies, Synflow (+ 0.87\%) and NASWOT (+3.01\%). With 20 architectures for DPS and 40 queries (total of 60) to retrieve the top architectures during RE, Eproxy+DPS achieves better results than existing SoTA predictor-based NAS WeakNAS with 100 queries (+0.23\%). Furthermore, we explore the 70 neighbors of the top architectures (a total of 150 queries) and find architectures with an average of 94.23\% accuracy. Note that Semi-NAS with 1000 queries can only reach 94.01\%. 
On \textbf{NAS-Bench-201}, we perform the DPS on the CIFAR-10 dataset, and the found proxy is directly transferred to CIFAR-100 and Tiny-ImageNet. We compare with MCTS~\citep{wang2019alphax}, LaNAS~\citep{wang2021sample}, WeakNAS~\citep{wu2021stronger}. In Table~\ref{table:end2end}, we show that Eproxy+DPS can find optimal global architectures within the RE search history. Compared to RE, which directly queries the benchmark, our approach reduced 7x/32x/9x query times on three datasets. Compared to predictor-based NAS, Eproxy+DPS also requires fewer queries to discover the optimal architectures. Our results offer an exciting yet promising direction besides pure predictor-based NAS.

\textbf{Open DARTS-ImageNet search space}
~\label{sec:imagenet} On DARTS search space~\citep{liu2018darts}, we perform the end-to-end search on ImageNet-1k~\citep{deng2009imagenet} dataset. The networks' depth (number of micro-searching blocks) is 14. The input channel number is 48, and architectures are with FLOPs between 500M to 600M. 
We utilize the 20 samples from the NDS-DARTS search space (not the same search space as the target) and conduct DPS on CIFAR-10 for 200 epochs in a GPU hour. 
Then we perform the NAS by adopting regularized evolutionary algorithm with the loss of the zero-cost proxy as the fitness function in 0.4 GPU hour. We compare our method with (a) existing works on the DARTS search space~\cite{liu2018darts,xie2018snas,dong2019searching,chen2019progressive,xu2019pc,chen2021neural} and (b) works on the similar search spaces~\cite{zoph2018learning,real2019regularized,liu2018progressive,cai2018proxylessnas}. The results are shown in Table~\ref{table:GenNAS_NAS_Imagenet}. Eproxy achieves a top-1/5 test error of 25.2/8.1 using Eproxy with only 0.5 GPU hours for NAS. With DPS, Eproxy explores the architecture with 24.4\%/7.3\% as a top-1/top5 test error.
Eproxy+DPS significantly outperforms existing NAS on CIFAR-10, such as PC-DARTS, and achieves a comparable result with NAS on ImageNet, demonstrating Eproxy and DPS's efficiency. By utilizing the existing performance of architectures on another dataset/search space, DPS shows the transferability between tasks and search spaces. 

\section{Conclusion}
In this work, we proposed Eproxy that utilizes a self-supervised few-shot regression task within near-zero cost. The Eproxy is benefited from the barrier layer that significantly improves the complexity of the proxy task. To overcome the drawbacks of current efficient proxies that are not adaptive to various tasks/search spaces, we proposed DPS incorporating various settings and hyperparameters in a proxy search space and leveraging REA to conduct efficient exploration. Our experiments on numerous NAS benchmarks demonstrate that Eproxy is a robust, efficient proxy. Moreover, with the help of DPS, Eproxy achieves state-of-the-art results and outperforms existing state-of-the-art efficient proxies, early stopping methods and predictor-based NAS. Our work significantly ameliorates the inconsistency of efficient proxies and sets up a series of solid baselines while pointing out a novel direction for the NAS community.


\bibliography{iclr2023}
\bibliographystyle{plainnat}

\appendix
\section{Appendix}
\subsection{Experiment Setup}
 \textbf{Eproxy} The learning rate is $1.0$, and the weight decay is $1\mathrm{e}{-5}$. Each architecture is trained for ten iterations with 16 images randomly sampled from the CIFAR-10 training set as a mini-batch (tiny dataset). The SGD optimizer is used for training. 
 
 \textbf{DPS} The total evolution cycle is 200. The number of architectures sampled for ranking is 20. The population size is 40. The sample size is 10. The mutation rate is 0.2.

\subsection{GPU Benchmark}
We benchmark the average evaluation time for architecture with Eproxy and GPU utilization on different search spaces (shown in Table~\ref{table: gpubenchmark}). For DPS, it's straightforward to estimate the total time. For example, if we conduct DPS on NDS-DARTS search space with 20 architectures to get each proxy's ranking correlation and 200 total evolution cycles, the time is $\sim 20\times 200\times 0.72=2880$ seconds. All experiments are done on a single A6000 GPU.

\begin{table}[t]
\centering
\scriptsize
\begin{tabular}{lccccc}\toprule
Search space        & NB101  & NB201        & DARTS     & DARTS-fix-w-d         & Amoeba       \\\midrule
Avg. Eval. Time (ms) &    414.1    &     324.0         &      719.2     &           1198.3            &      1191.3        \\\midrule
GPU Util. (MB)      &    4137    &       1603       &     3221      &               2275        &      3365        \\\midrule
Search space        & ENAS   & ENAS-fix-w-d & NASNet    & PNAS                  & PNAS-fix-w-d \\\midrule
Avg. Eval. Time (ms) &   908.2     &      1408.2        &      878.7     &        1041.4               &        1824.7      \\\midrule
GPU Util. (MB)      &    3245    &     2577         &      3129     &               3391        &           3447   \\\midrule
Search space        & ResNet & ResNeXt-A    & ResNeXt-B & NAS-Bench-Trans-Micro & NAS-Bench-MR \\\midrule
Avg. Eval. Time (ms) &    242.3      &     314.5         &      298.7     &      355.2                 &        1011.9      \\\midrule
GPU Util. (MB)      &    2765    &        2423      &     2777      &          2081             &      4229       \\\bottomrule
\end{tabular}\caption{Average time for evaluating an architecture with Eproxy in the target search space and Maximum GPU utilization. The results suggest that Eproxy is efficient and computation-friendly. }\label{table: gpubenchmark}
\end{table}

\subsection{Search Spaces}
\textbf{NAS-Bench-101}~\citep{ying2019bench}: 423K CNN architectures are trained on CIFAR-10 dataset. 

\textbf{NAS-Bench-201}~\citep{dong2020nasbench201}: 15625 CNN architectures are trained on CIFAR-10/CIFAR-100/TinyImageNet.

\textbf{NDS dataset}~\citep{radosavovic2020designing}: \textbf{DARTS}: A DARTS~\citep{liu2018darts} style search space including 5000 sampled architectures trained on CIFAR-10.
\textbf{DARTS-fix\_w\_d}: A DARTS style search space with fixed width and depth including 5000 sampled architectures trained on CIFAR-10.
\textbf{AmoebaNet}: An AmoebaNet~\citep{real2019regularized} style search space including 4983 sampled architectures trained on CIFAR-10.
\textbf{ENAS}: An ENAS~\citep{pham2018efficient} style search space including 4999 sampled architectures trained on CIFAR-10.
\textbf{ENAS-fix\_w\_d}: An ENAS style search space with fixed width and depth including 5000 sampled architectures trained on CIFAR-10.
\textbf{NASNet}: A NASNet~\citep{zoph2018learning} style search space including 4846 sampled architectures trained on CIFAR-10.
\textbf{PNAS}: A PNAS~\citep{liu2018progressive} style search space including 4999 sampled architectures trained on CIFAR-10.
\textbf{PNAS-fix\_w\_d}: A PNAS style search space with fixed width and depth including 4559 sampled architectures trained on CIFAR-10.
\textbf{ResNet}: A ResNet~\citep{he2016deep} style search space including 25000 sampled architectures trained on CIFAR-10.
\textbf{ResNeXt-A}: A ResNeXt~\cite{xie2017aggregated} style search space including 24999 sampled architectures trained on CIFAR-10.
\textbf{ResNeXt-B}: Another ResNeXt style search space including 25508 sampled architectures trained on CIFAR-10.
\textbf{DARTS\_in}: A DARTS style search space including 121 sampled architectures trained on ImageNet-1k.
\textbf{DARTS-fix\_w\_d-in}: A DARTS style search space with fixed width and depth including 499 sampled architectures trained on ImageNet-1k.
\textbf{Amoeba\_in}: An AmoebaNet style search space including 124 sampled architectures trained on ImageNet-1k.
\textbf{ENAS\_in}: A ENAS style search space including 117 sampled architectures trained on ImageNet-1k.
\textbf{NASNet\_in}: A NASNet style search space including 122 sampled architectures trained on ImageNet-1k.
\textbf{PNAS\_in}: A PNAS style search space including 119 sampled architectures trained on ImageNet-1k.
\textbf{ResNeXt-A\_in}: A ResNeXt style search space including 130 sampled architectures trained on ImageNet-1k.
\textbf{ResNeXt-B\_in}: Another ResNeXt style search space including sampled 164 architectures trained on ImageNet-1k.

\textbf{NAS-Bench-Trans-Micro}~\cite{duan2021transnas}: A NAS-Bench-201 style search space including 4096 architectures trained on 7 different tasks on the subsets of Taskonomy dataset~\citep{zamir2018taskonomy}. Tasks including: 
\textbf{Object Classification} for 75 classes of objects. 
\textbf{Scene Classification} for 47 classes of scenes. 
\textbf{Room Layout}  for estimating and
aligning a 3D bounding box by utilizing a 9-dimension vector.
\textbf{Jigsaw Content Prediction} by dividing the input image into 9 patches and shuffling according to one of 1000 preset permutations.
\textbf{Semantic Segmentation} for  17 semantic classes.
\textbf{Autoencoding} for reconstructing the input images.

\textbf{NAS-Bench-MR}~\citep{ding2021learning}: A complex search space for multi-resolution networks including 2507 trained architectures on 9 different tasks. Tasks including:  
\textbf{ImageNet-50-1000
(Cls-A)} with 50 classes and 1000 samples from each class from ImageNet-1k. 
\textbf{ImageNet-50-100
(Cls-B)} with 50 classes and 100 samples from each class from ImageNet-1k. 
\textbf{ImageNet-10-1000
(Cls-A)} with 10 classes and 1000 samples from each class from ImageNet-1k. 
\textbf{ImageNet-10c} same as Cls-A but architectures are trained for 10 epochs.
\textbf{Seg} for Cityscapes dataset~\citep{cordts2016cityscapes}.
\textbf{Seg-4x} for Cityscapes dataset with 4x downsampled resolution.
\textbf{3dDet} on KITTI dataset~\citep{Geiger2012CVPR}.
\textbf{Video} for HMDB51 dataset~\cite{kuehne2011hmdb}.
\textbf{Video-p} for HMDB51 but architectures are pretrained with ImageNet-50-1000.

\subsection{Searched Architectures}
\begin{figure}[h]
\vspace{-16pt}
     \centering
          \begin{subfigure}[b]{0.49\textwidth}
         \centering
         \includegraphics[width=1.\textwidth]{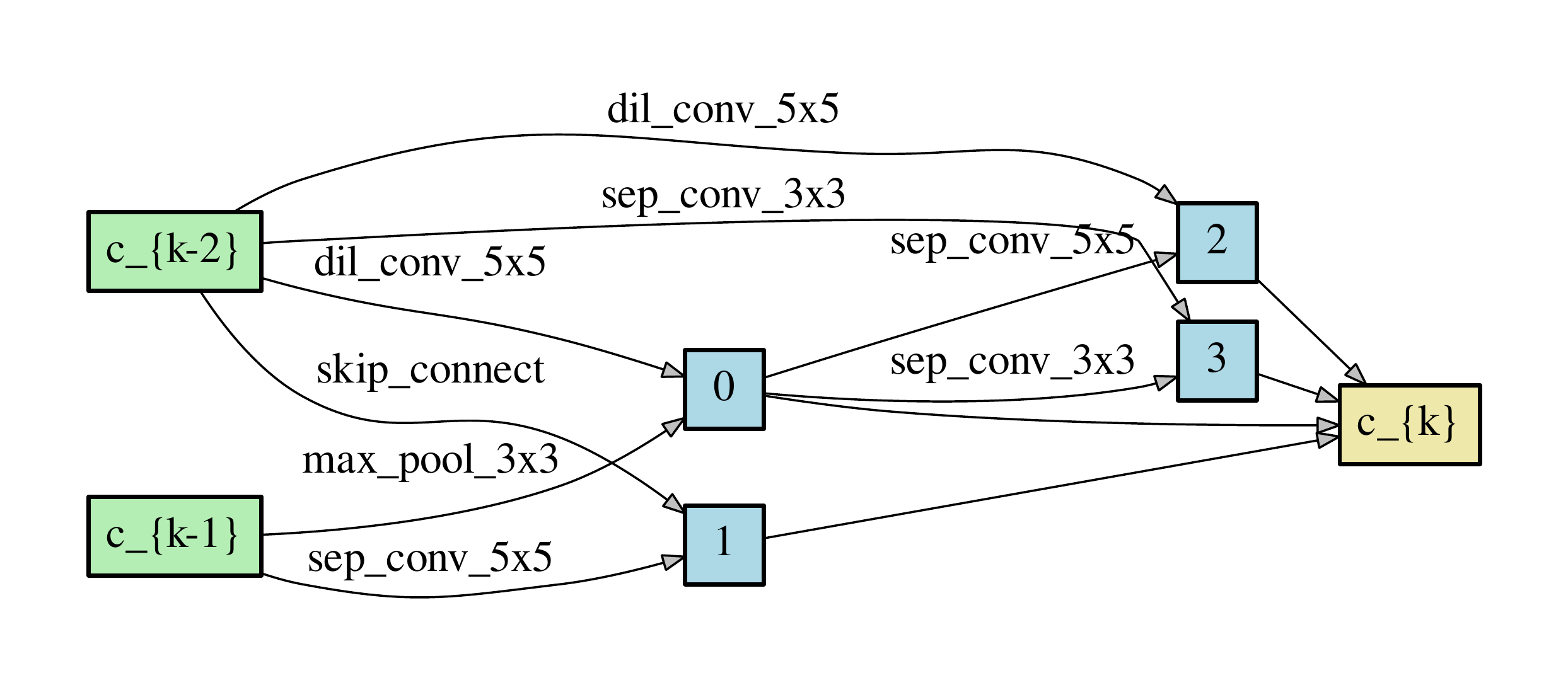}
         \caption{Eproxy Normal Cell}
     \end{subfigure}
     \hfill
     \begin{subfigure}[b]{0.49\textwidth}
         \centering
         \includegraphics[width=\textwidth]{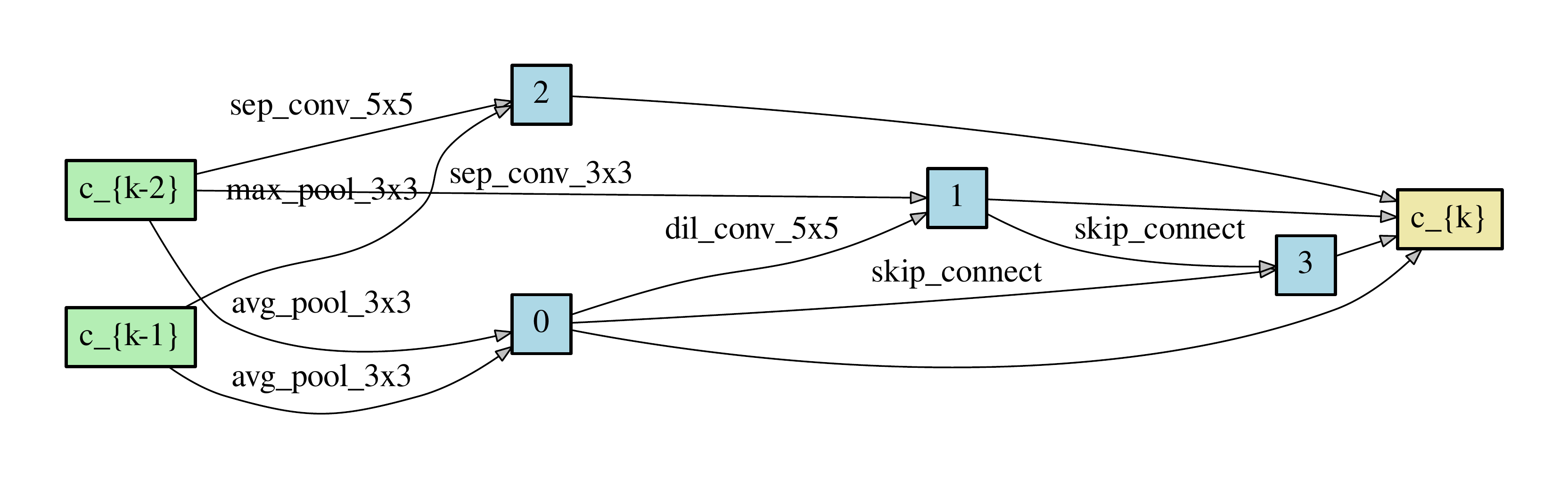}
         \caption{Eproxy Reduction Cell}
     \end{subfigure}
     \begin{subfigure}[b]{0.49\textwidth}
         \centering
         \includegraphics[width=1.\textwidth]{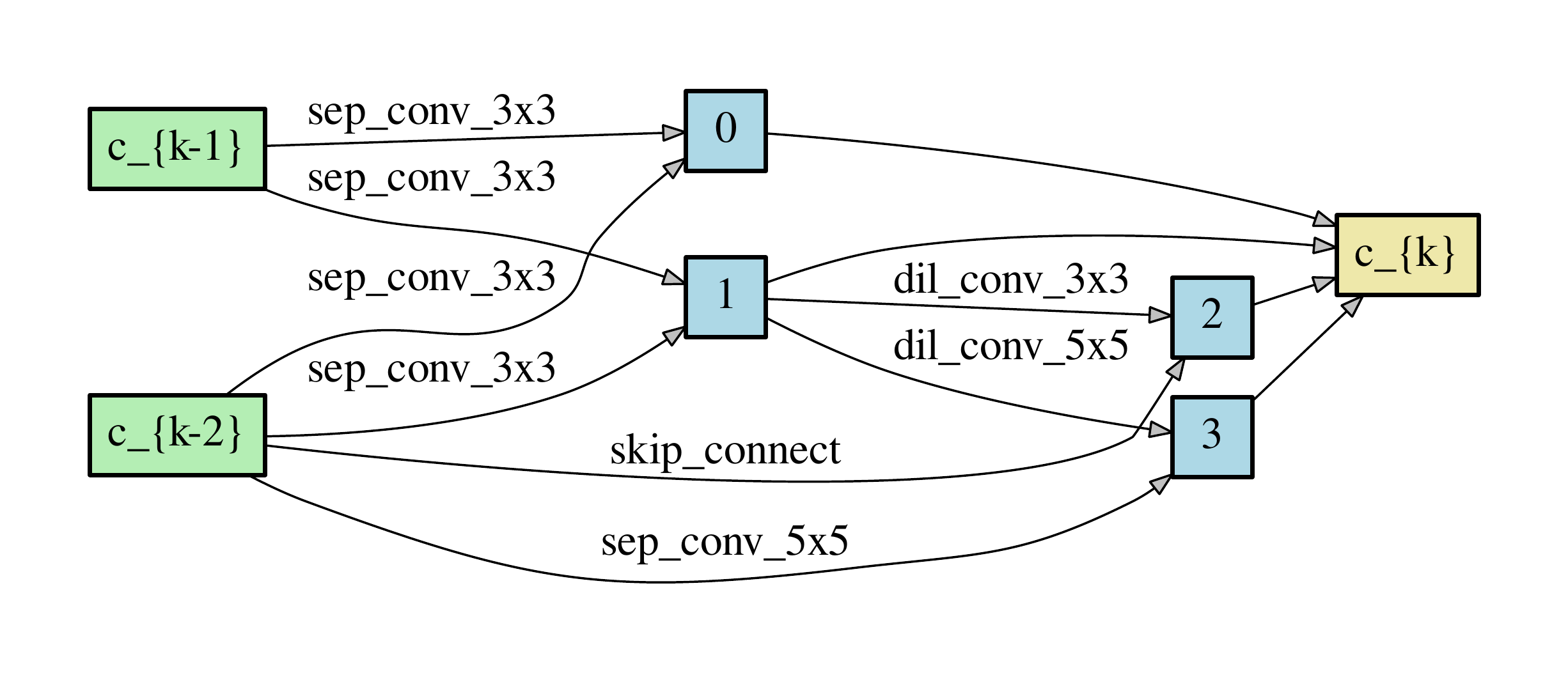}
         \caption{Eproxy+DPS Normal Cell}
     \end{subfigure}
     \hfill
     \begin{subfigure}[b]{0.49\textwidth}
         \centering
         \includegraphics[width=\textwidth]{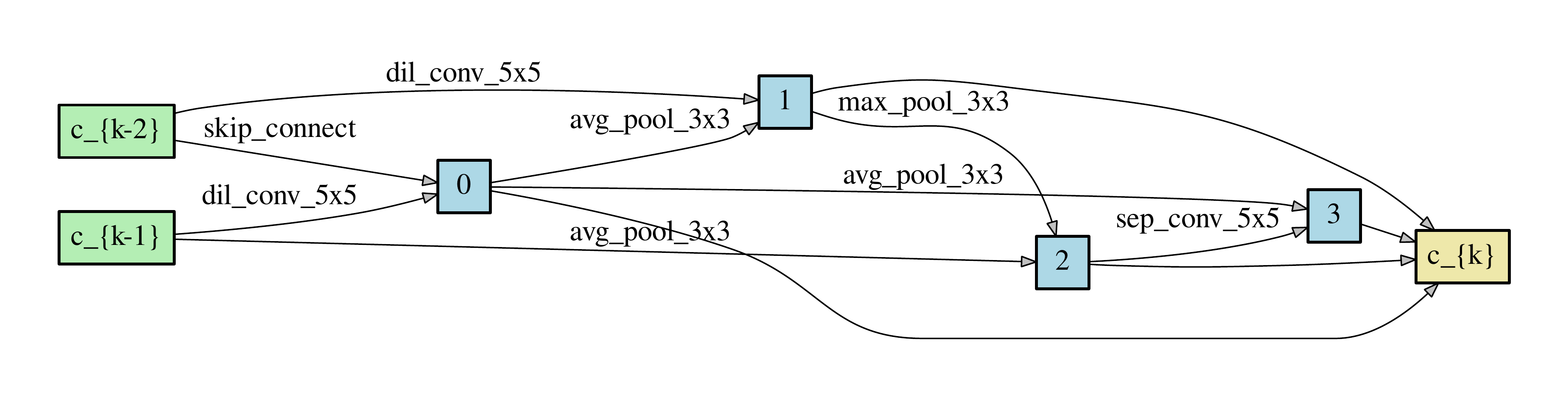}
         \caption{Eproxy+DPS Reduction Cell}
     \end{subfigure}
     \hfill
        \caption{Visualize the architecture found by Eproxy and Eproxy+DPS on ImageNet-DARTS search space.}
        \label{fig:arch}
        \vspace{-8pt}
\end{figure}

The searched architectures for DARTS-ImageNet search space are shown in Fig~\ref{fig:arch}.

\subsection{Pseudo Code for Eproxy}
\begin{lstlisting}[language=Python, label={list:eproxy}, caption=Pseudo PyTorch-sytle~\cite{paszke2019pytorch} code for Eproxy.]{Name} 

def Eproxy(model, barrier, img, label, t_iter = 10):
    # img shape: B, C = 3, W_in, H_in
    # label shape: B,  C_out, W_out, H_out
    optimizer = torch.optim.SGD(model.parameters(),
                                lr=1.0,
                                momentum=0.9,
                                weight_decay=4e-5)
    for i in range(t_iter):
        output_mid = model(img) # B, C_mid, W_out, H_out
        output = barrier(output_mid) # B, C_out, W_out, H_out
        loss = ((output - label)**2).mean()
        optimizer.zero_grad()
        loss_m.backward()
        optimizer.step()
    return loss
\end{lstlisting}

\subsection{Pseudo Code for DPS}
\begin{lstlisting}[language=Python, label={list:dps}, caption=Pseudo PyTorch-sytle code for DPS.]{Name} 

def DPS(archs_accs, cycle, population = 40, sample = 10, mutation_rate = 0.2):
    # len(archs_accs): 20
    # config: including lr, channel number, feature combination, etc.
    config_history = []
    rea = REAEngine(population, sample, mutation_rate)
    # generate initial pool
    for _ in range(population):
        config = rea.get_random_config()
        rank = rea.get_rank(config, archs_accs)
        config_history.append({'config': config, 'rank': rank})
    # evolution
    for _ in range(cycle):
        new_config = rea.get_mutate_config()
        rank = rea.get_rank(new_config, archs_accs)
        config_history.append({'config': new_config, 'rank': rank})
        # rea.get_config_pool().size(): 40
        # rea.get_config_pool_rank().max(): the proxy in the pool with highest ranking correlation on the archs_accs set)
    return config_history
\end{lstlisting}

\subsection{More Loss landscapes}\label{sec: moreloss}
\begin{figure}[b]
\vspace{-8pt}
     \centering
     \begin{subfigure}[b]{0.49\textwidth}
         \centering
         \includegraphics[width=\textwidth]{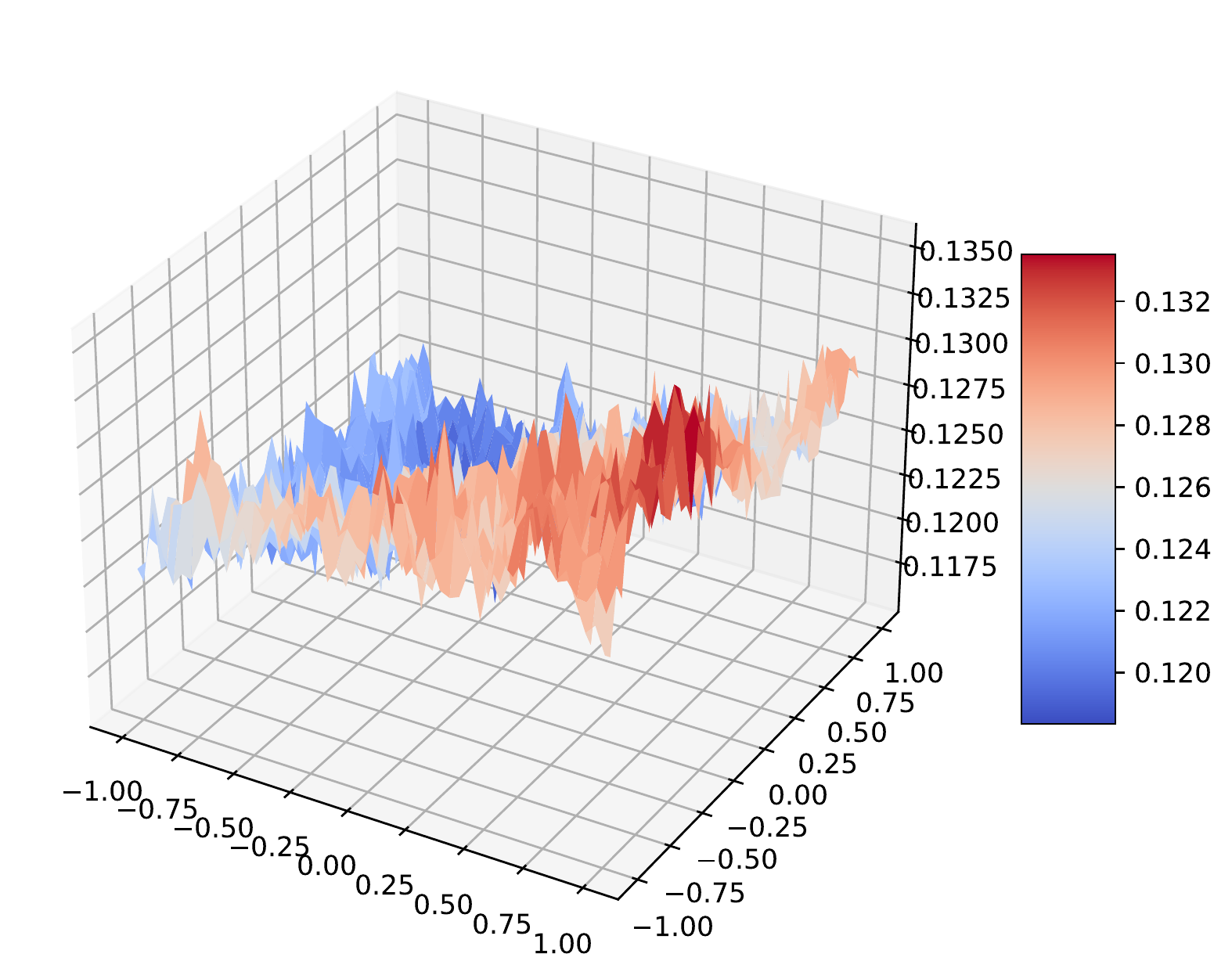}
         \caption{Best model with the barrier.}
         \label{fig:barr_best}
     \end{subfigure}
     \hfill
     \begin{subfigure}[b]{0.49\textwidth}
         \centering
         \includegraphics[width=\textwidth]{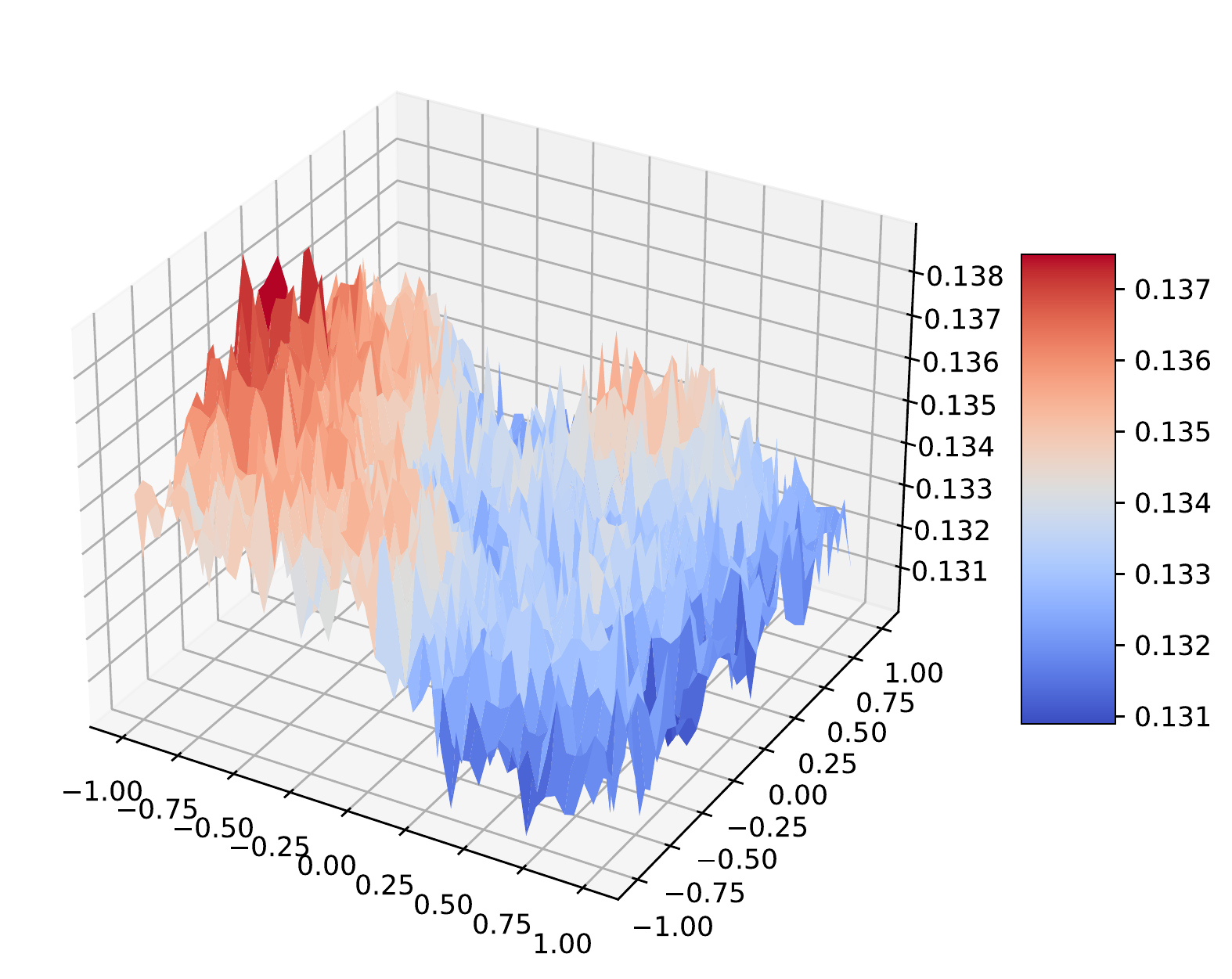}
         \caption{Worst model with the barrier.}
         \label{fig:barr_worst}
     \end{subfigure}
        \caption{The loss surfaces of best and worst model from NAS-Bench-101 regression task with barrier.}
        \label{fig:barr}

     \centering
     \begin{subfigure}[b]{0.49\textwidth}
         \centering
         \includegraphics[width=\textwidth]{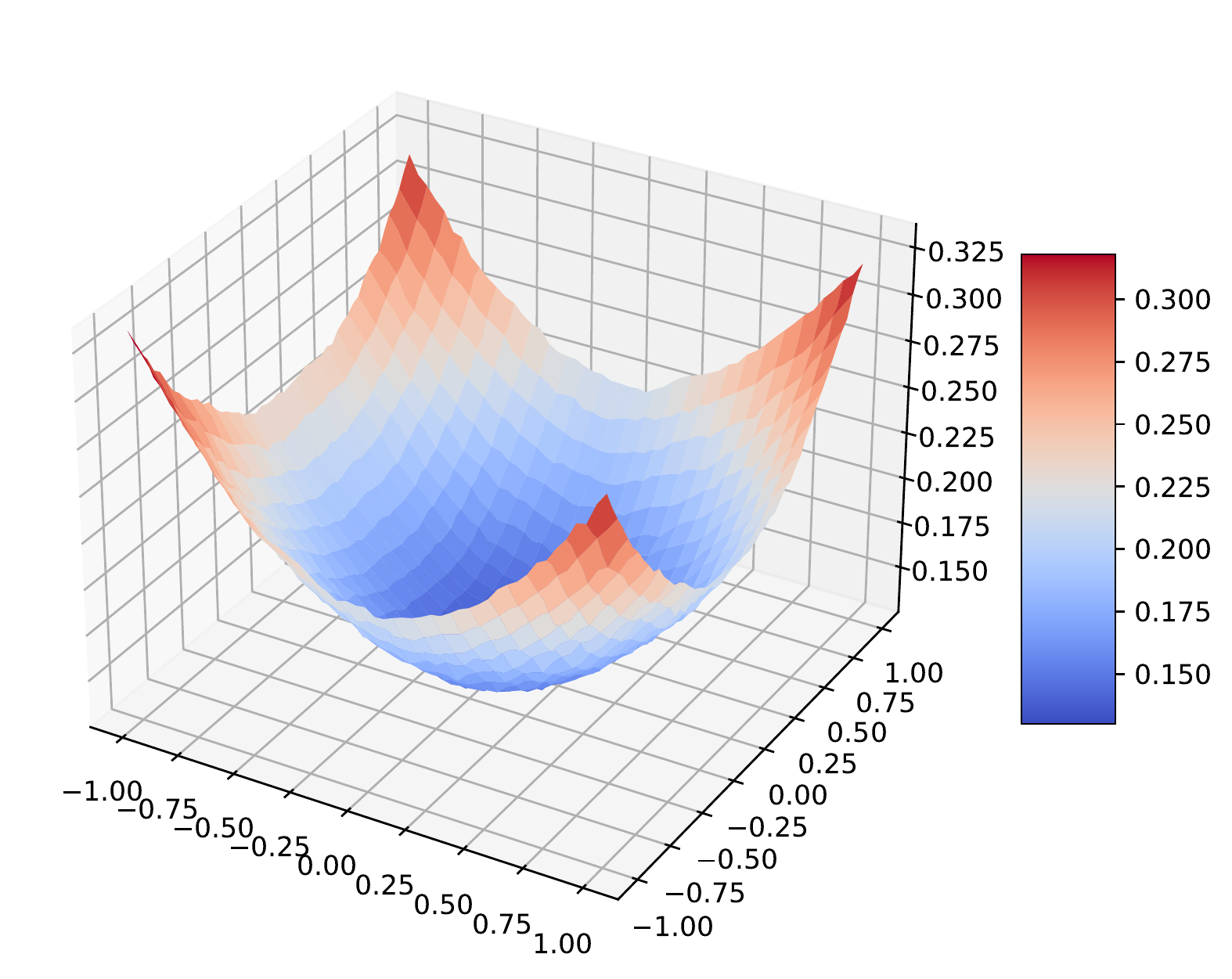}
         \caption{ Best model without barrier.}
         \label{fig:nobarr_best}
     \end{subfigure}
     \hfill
     \begin{subfigure}[b]{0.49\textwidth}
         \centering
         \includegraphics[width=\textwidth]{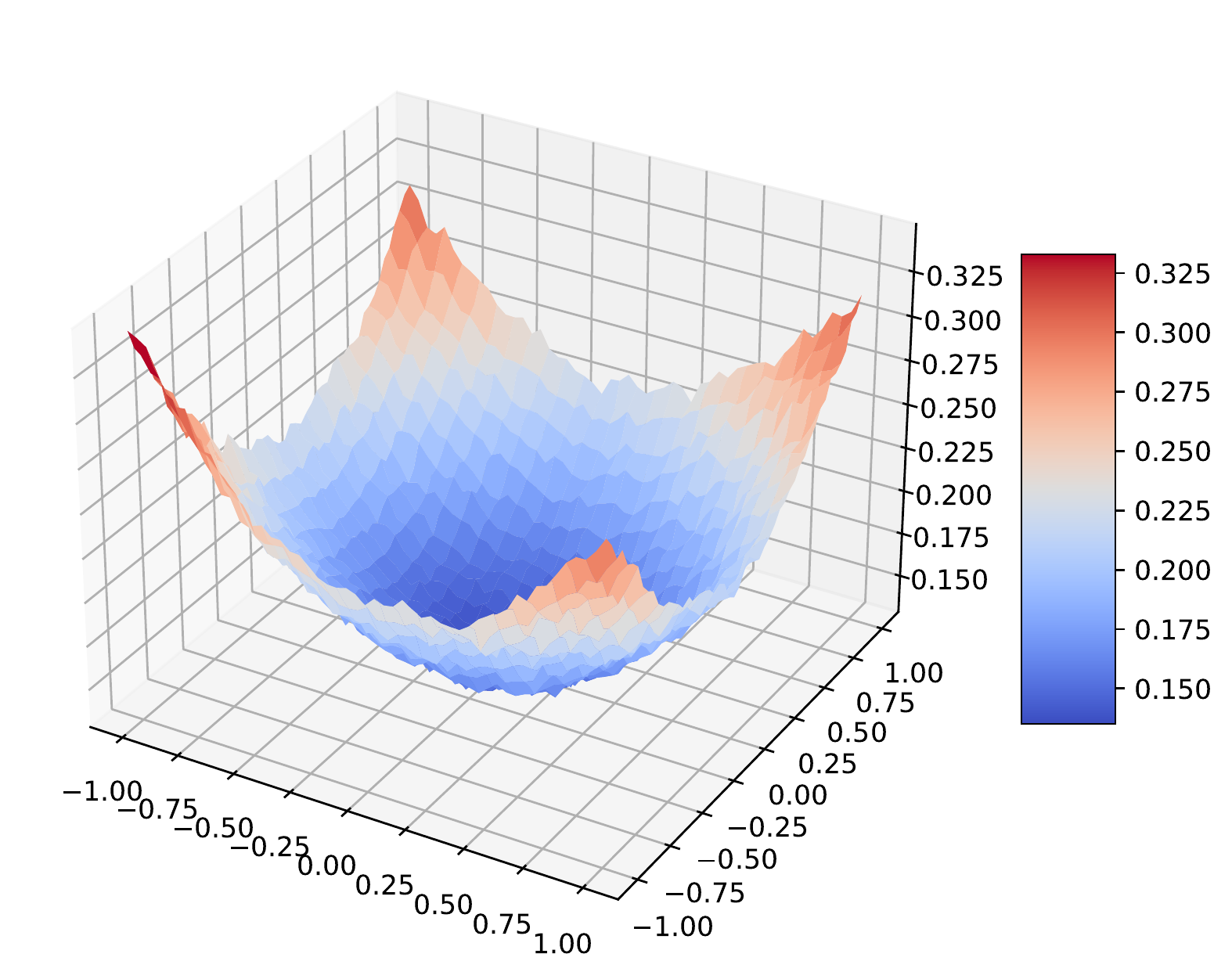}
         \caption{ Worst model without barrier.}
         \label{fig:nobarr_worst}
     \end{subfigure}
        \caption{The loss surfaces of best and worst model from NAS-Bench-101 regression task without barrier.}
        \label{fig:nobarr}
        \vspace{-8pt}
        
\end{figure}

We listed more loss landscapes from the best and the worst models in NAS-Bench-101~\citep{ying2019bench} search space on our proxy task, either with or without the barrier in Fig.~\ref{fig:barr} and Fig.~\ref{fig:nobarr}. From Fig.~\ref{fig:nobarr}, we can observe that the best model has a much smoother loss surface than the worst model. From Fig.~\ref{fig:barr}, we can observe that the best model's can achieve lower loss compared to worst model even though the loss surface is sophisticated. Besides, the loss surfaces are significantly different which means the optimization directions for both models are distinctive. We can also observe from Fig.~\ref{fig:nobarr} that the best and worst models have similar convexity and shape, which makes the proxy task produce a much worse ranking correlation score compared with the proxy task that uses the barrier.

\subsection{Limitations}

1. Though empirical results strongly support Eproxy and DPS, there is no strict mathematical proof of the upper bound of the similarity between a few-shot proxy task and a large-scale task.
2. Our experiments are limited to Computer Vision tasks. It is unknown whether the Eproxy can be extended to Natural Language Processing tasks~\citep{vaswani2017attention,hochreiter1997long,schuster1997bidirectional,devlin2018bert}. 

\end{document}

%% file: math_commands.tex
\usepackage{amsmath,amsfonts,bm}

\def\eqref#1{equation~\ref{#1}}

\def\1{\bm{1}}

\DeclareMathAlphabet{\mathsfit}{\encodingdefault}{\sfdefault}{m}{sl}
\SetMathAlphabet{\mathsfit}{bold}{\encodingdefault}{\sfdefault}{bx}{n}

